\pdfoutput=1
\documentclass{article}
\usepackage{times}
\usepackage{cite}
\usepackage{epsfig}
\usepackage{graphicx}
\usepackage{amsmath}
\usepackage{import}
\usepackage{amssymb}
\usepackage{subcaption}
\usepackage[subtle]{savetrees}

\newcommand{\comment}[1]{\textcolor{blue}{}}

\usepackage[breaklinks=true,bookmarks=false]{hyperref}

\usepackage{microtype}
\usepackage{booktabs} 

\usepackage[accepted]{icmlw2019generalization}

\icmltitlerunning{Visualizing How Embeddings Generalize}

\begin{document}

\twocolumn[
\icmltitle{Visualizing How Embeddings Generalize}

\icmlsetsymbol{equal}{*}

\begin{icmlauthorlist}
\icmlauthor{Xiaotong Liu}{gwu}
\icmlauthor{Hong Xuan}{gwu}
\icmlauthor{Zeyu Zhang}{gwu}
\icmlauthor{Abby Stylianou}{gwu}
\icmlauthor{Robert Pless}{gwu}
\icmlaffiliation{gwu}{Department of Computer science, The George Washington University,Washington, DC, USA}
\icmlcorrespondingauthor{Xiaotong Liu}{liuxiaotong2017@gwu.edu}
\icmlcorrespondingauthor{Robert Pless}{pless@gwu.edu}
\end{icmlauthorlist}

\icmlkeywords{Machine Learning, ICML}

\vskip 0.3in
]



\printAffiliationsAndNotice{\icmlEqualContribution} 

\begin{abstract}
Deep metric learning is often used to learn an embedding function that captures the semantic differences within a dataset. A key factor in many problem domains is how this embedding generalizes to new classes of data. In observing many triplet selection strategies for Metric Learning, we find that the best performance consistently arises from approaches that focus on a few, well selected triplets. We introduce visualization tools to illustrate how an embedding generalizes beyond measuring accuracy on validation data, and we illustrate the behavior of a range of triplet selection strategies.
\end{abstract}

\section{Introduction}
\label{submission}

Deep Metric Learning works to find embeddings that map semantically similar points to be nearby in an embedding space. Many imaging tasks start with datasets that include many classes and many images per class, and embeddings are trained to map images from the same class to be near each other and far from images of different classes. A key area of research is finding approaches that learn embeddings that satisfy this constraint and also generalize to new data so that this constraint also holds when new image classes are introduced.

\begin{figure}[t]
    \centering
    \begin{subfigure}[]{\columnwidth}
            \centering
            \includegraphics[width=.32\columnwidth]{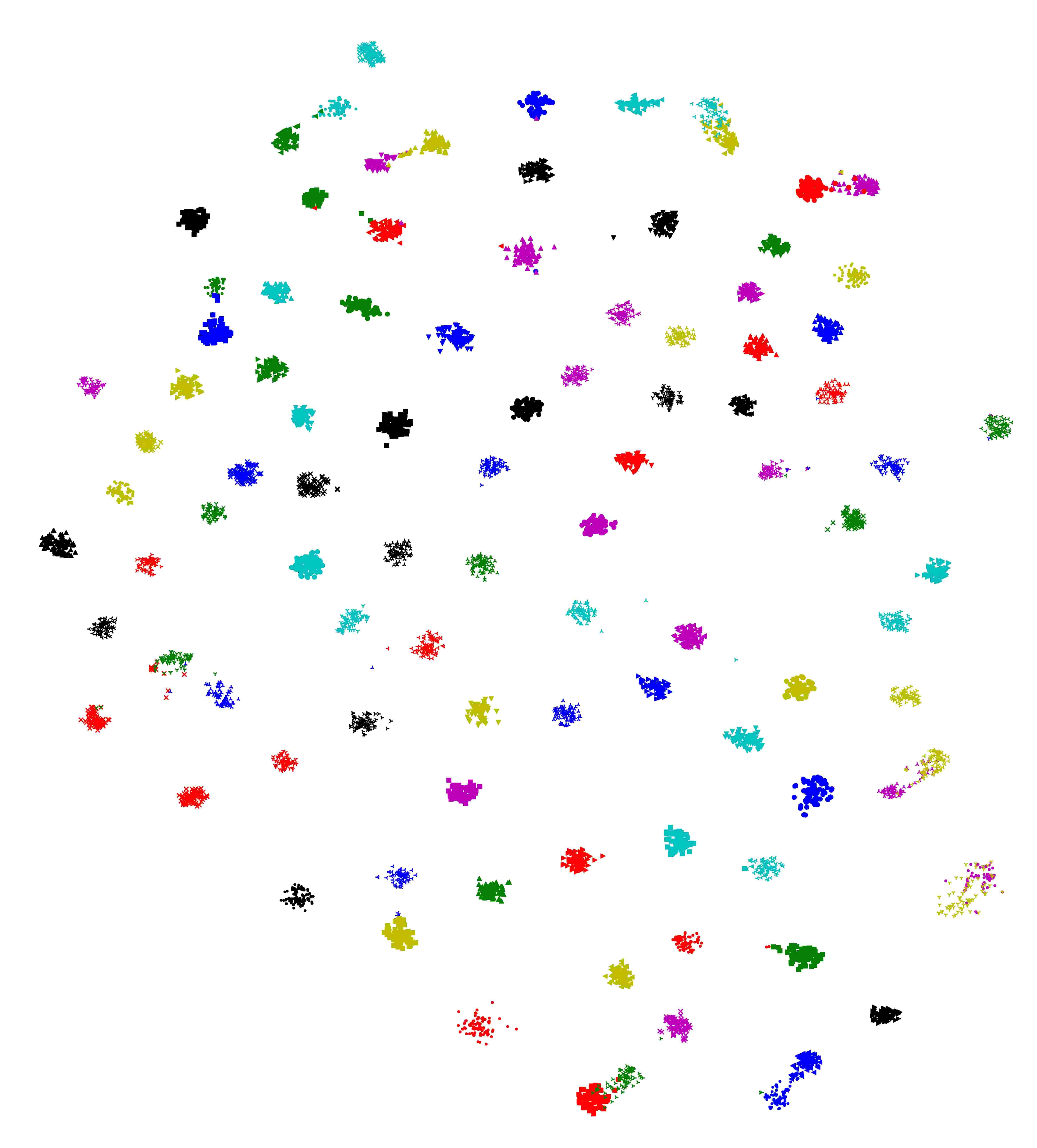}
            \includegraphics[width=.32\columnwidth]{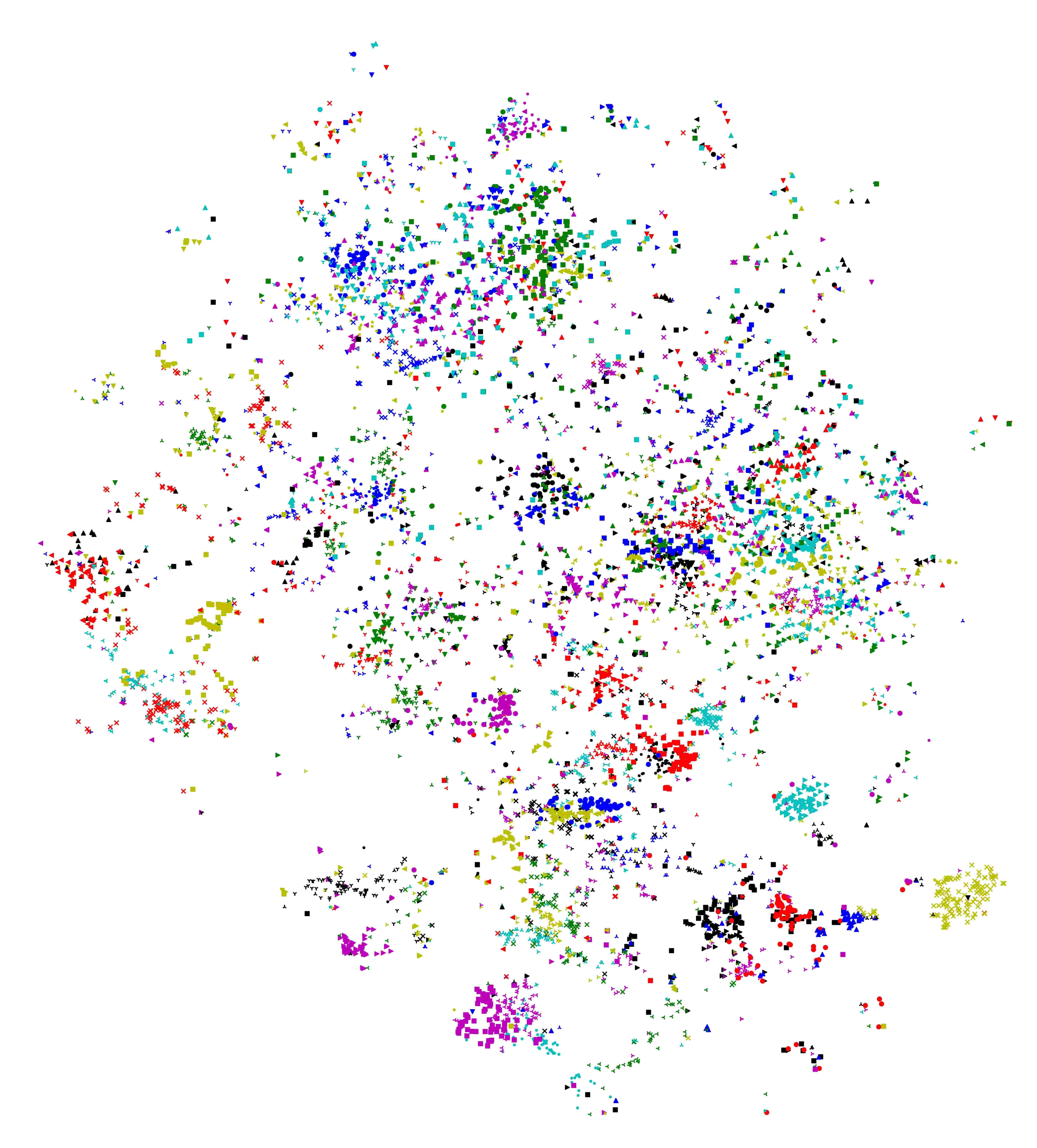}
            \includegraphics[width=.32\columnwidth]{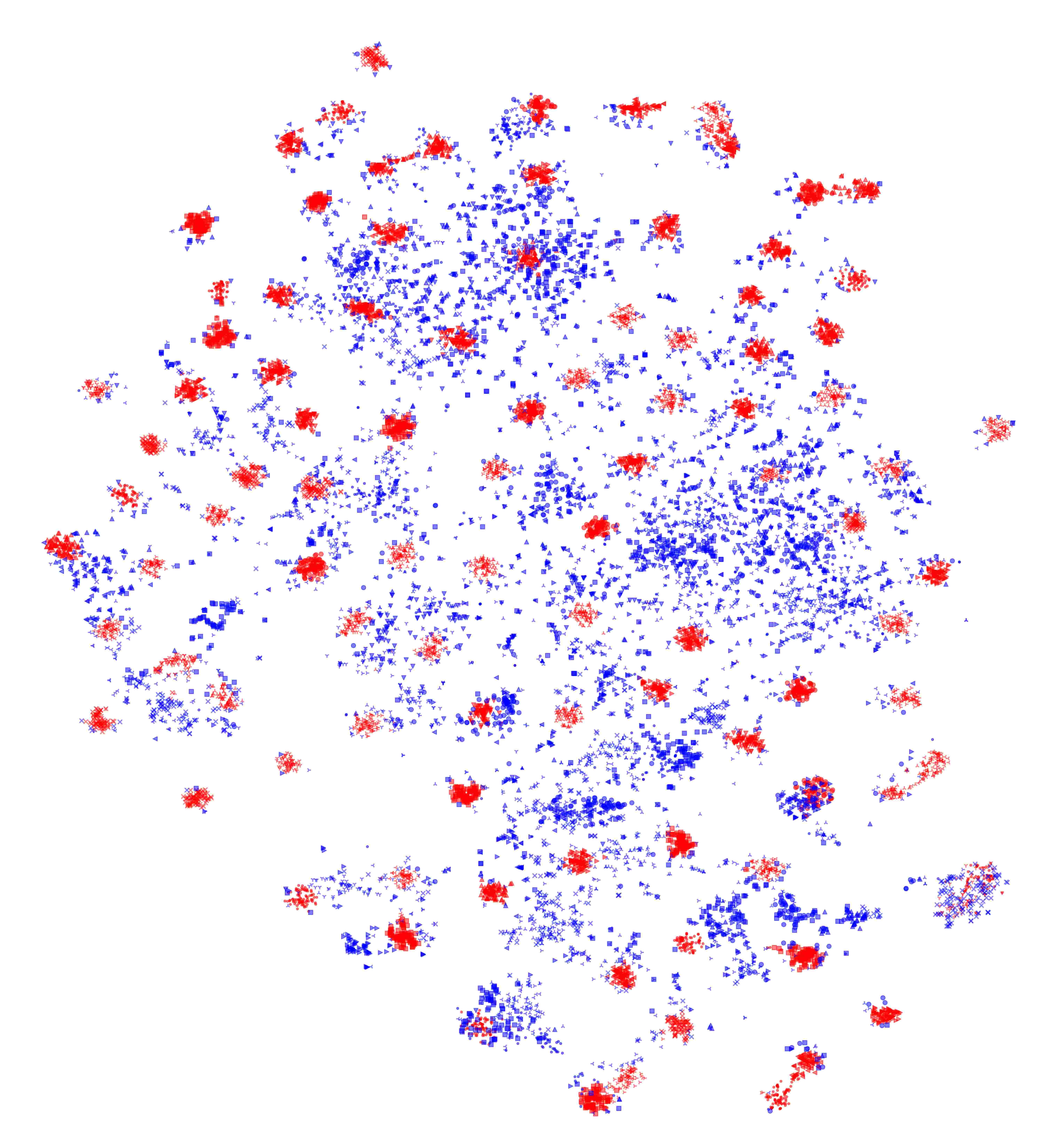}
            \caption{\label{fig:front_page_tsne}}
    \end{subfigure}
    \\
    \begin{subfigure}[]{\columnwidth}
        \centering
        \includegraphics[width=.49\columnwidth]{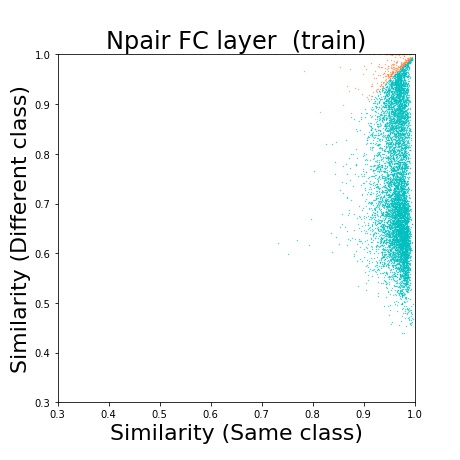}
        \includegraphics[width=.49\columnwidth]{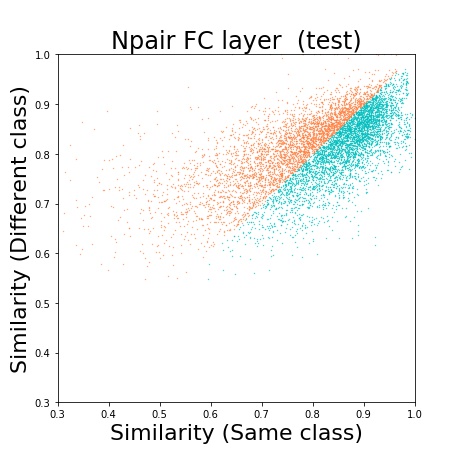}
        \caption{\label{fig:scatter}}
	\end{subfigure}
        \caption{We introduce several visualizations to characterize how embeddings generalize to new data.  For features extracted from the fully connected (FC) layer of a network trained on the CAR dataset with N-Pairs loss, (a) shows a joint t-SNE embedding showing (top-left) training data, (top-center) test data, and (top-right) both together; collectively, these show how the test data is embedded relative to the training data.  At a finer scale, scatter plots showing the distance to the most similar example from the same and different classes in (bottom-left) training and (bottom-right) testing data, highlight that an embedding trained by N-Pairs may fail to generalize to new classes because new class images may not be embedded to similar locations.}
\label{fig:new_front_page}
\end{figure}

A popular approach to learning these embeddings is based on triplets of data: an anchor, another example from the same class, and an example from a different class. A deep convolutional neural network is then optimized via a loss function that encourages the anchor image to be closer to the positive example than the negative example. While there is a large and growing literature of variations to this idea, a recent trend shows that approaches that are more selective about which triplets are considered in the training tend to generalize better.

We consider a set of approaches to selecting triplets from a batch of training data. These approaches are described in Section~\ref{sec:triplets}, and range from considering all possible triplets in a single batch to considering only a single, carefully selected triplet per example. We found it surprising that approaches that consider a single more carefully selected triplet per training example yield superior performance in generalizing to new classes. This was surprising because, compared to approaches that consider all possible triplets, these approaches seemingly lose much of the efficiency of optimizing based on the large numbers of possible triplets in a batch.

Furthermore, beyond reporting results on validation data that include new classes, there is a lack of tools to understand this generalization. Therefore, we contribute in this paper simple visualizations to help highlight how embedding approaches generalize to new data. We also include a discussion about what these visualizations tell us about why different embedding approaches are more or less able to give good results when generalizing to new data classes.

\section{Methods}
We train a selection of recently introduced approaches for Deep Metric Learning. In this section, we define the specifics of how we train these methods, and the details of the visualization approaches we use to compare them.


\subsection{Embedding Methods}
\label{sec:triplets}
Triplet loss is trained with triplets of images, $(x_a,x_p,x_n)$, where $x_a$ is an anchor image, $x_p$ is a positive image of the same class as the anchor, and $x_n$ is a negative image of a different class, and the convolutional neural network, $f(\cdot)$, embeds the images on a unit sphere, $(f(x_a), f(x_p), f(x_n))$. The target is to learn an embedding such that the anchor-positive pair are closer together than the anchor-negative pair. 

In this work we compare different approaches to constructing triplets within a batch. The first set of approaches start with each example in a batch and construct all or many possible triplets for that example. Batch All~\cite{HermansBeyer2017Arxiv} considers all possible triplets in a batch, and N-pair loss~\cite{Npairs} adds more negative examples into triplets and turns the triplet into an N-tuple, $(x_a,x_p,x_{n_1},...,x_{n_i})$. The second set of approaches considers each example, and constructs a single, carefully selected triplet for that example.  Semi-Hard Negative~\cite{facenet} chooses a random anchor-positive pair and an anchor-negative pair that is farther than the anchor-positive pair, but within a margin so that its similarity is comparable. The recently introduced Easy Positive Semi-Hard Negative (EPSHN)~\cite{xuan2019improved} constructs a single triplet per image in a batch from its most similar same class example and a semi-hard negative example, and is the best performing single network (non-ensemble) approach to date. In all cases, we modify the standard triplet loss function from a margin based loss to an NCA loss~\cite{nca}, which we have found to give better performance.

\subsection{Network Details}
All tests are run on the PyTorch platform~\cite{pytorch}, using the ResNet18~\cite{resnet} architecture, pre-trained on ILSVRC 2012-CLS data~\cite{ILSVRC15}, to output a 64-dimensional representation. Training images are re-sized to 256 by 256 pixels. We adopt a standard data augmentation scheme (random horizontal flip and random crops padded by 10 pixels on each side). For pre-processing, we normalize the images using the channel means and standard deviations. All networks are trained using stochastic gradient descent (SGD). On all datasets we train using a batch size of 128. 

\subsection{Variations on t-SNE}
We introduce a flexible extension of t-SNE showing training and testing data jointly in this section, and use Dynamic t-SNE~\cite{dynamicTsne} to ``yoke'', or align, different t-SNE embeddings of the same data. Taken together, these t-SNE visualizations provide a intuitive view to compare and visualize high dimensional data. 

\paragraph{Joint Embedding}
t-SNE~\cite{maaten2008visualizing} is commonly used to show a 2D representation of high-dimensional data, such as the embedding results of a dataset like CAR~\cite{CAR196}. To explore how embeddings generalize, we suggest running t-SNE to find 2-D visualization for all points in \textit{both} training and testing dataset. This creates a unified embedding of both.  Figure~\ref{fig:front_page_tsne} visualizes this in 3 parts, showing the training data, color coded by class, the testing data, color coded by class, and the two overlayed, color coded by whether they come from training data (red) or testing data (blue).

\paragraph{Yoked t-SNE}
In order to directly compare between different t-SNE embeddings, we use the approach described in~\cite{dynamicTsne} to ``yoke'' the t-SNE embeddings of related high-D point sets together. For related embeddings (such as the embeddings of the CAR data set with two different networks), they suggest simultaneously optimizing the standard t-SNE on each one and including an alignment error term that penalizes the Euclidean distance between where the same point is mapped in each t-SNE embedding. By giving a very small weight to this alignment error, the t-SNE representation of each embedding remains similar, but when possible, the two embeddings are encouraged to place points in similar locations. All of the t-SNE plots in Figures~\ref{fig:new_front_page}~and~\ref{fig:full_page} have been aligned in this way.

\subsection{Same vs. Different Similarity Plots}
t-SNE is known to have limitations~\cite{wattenberg2016use} arising from mapping a high-dimensional space down to two dimensions, so we also propose to view the embedding that explicitly focuses on the exact similarity between a point and its closest same class and different class image in the original embedding space. Specifically, for all points in the dataset we create a scatter plot that shows the similarity to the closest same class image vs. the similarity to the closest different class image. Points that are below the $y=x$ diagonal, are closest to the same class, and (if used as a query) would be classified correctly with a nearest neighbor classifier. 

\begin{table}[t]
\setlength{\tabcolsep}{0.4em}
\begin{center}
\begin{tabular}{|c|c|c|c|c|}
\hline
Method & BatchAll & Npair & SHN & EPSHN \\
\hline
FC (train) 
 & 70.75 & 96.65 & 94.31 & 86.88\\
GAP (train) 
 & 73.69 & 85.47 & 91.93 & 84.79\\
FC (test) 
 & 33.72 & 53.13 & 60.61 & 73.22\\
GAP (test) 
 & 46.66 & 74.85 & 80.20 & 81.85\\

\hline
\end{tabular}
\end{center}
\caption{Recall@1 Performance on the CAR dataset}
\label{table:CAR}
\end{table}

\newcommand{\fullsizeplotwidth}{0.15\textwidth}
\newcommand{\fullsizeplotwidthdist}{0.3\textwidth}
\setlength{\tabcolsep}{1pt}
\renewcommand{\arraystretch}{.5}
\begin{figure*}
    \centering
    \begin{tabular}{ccccccc}
        \centering
        & \parbox{1.5cm}{\raggedright \centering t-SNE\\ (train)}
        & \parbox{1.5cm}{\raggedright \centering t-SNE\\ (test)} 
        & \parbox{1.5cm}{\raggedright \centering t-SNE\\ (combined)} 
        & \parbox{1.5cm}{\raggedright \centering scatter\\ (train)} 
        & \parbox{1.5cm}{\raggedright \centering scatter\\ (test)}\\
        
        \raisebox{.7cm}{\rotatebox{90}{Batch-All}} & 
        \includegraphics[width=\fullsizeplotwidth]{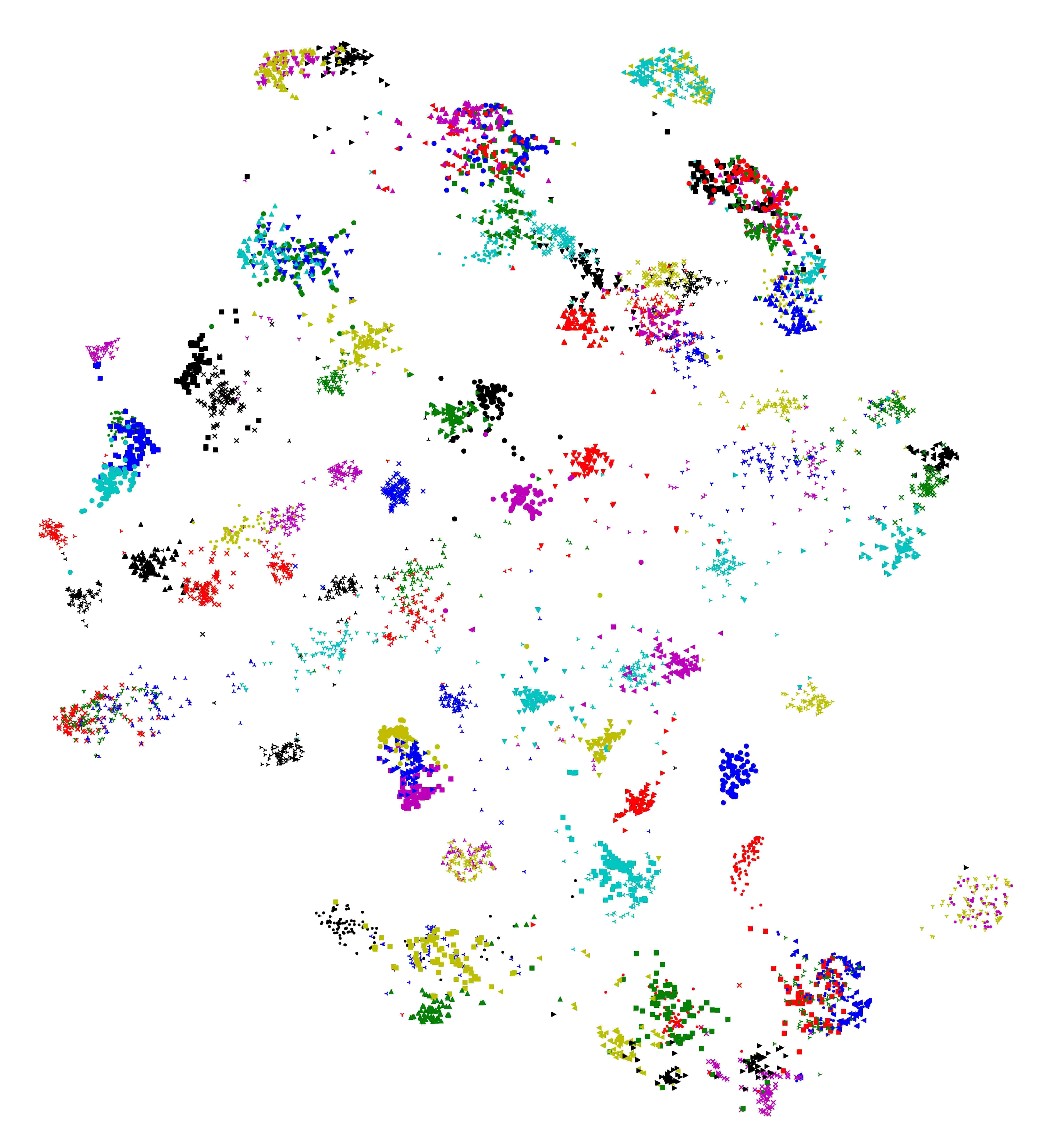} &
        \includegraphics[width=\fullsizeplotwidth]{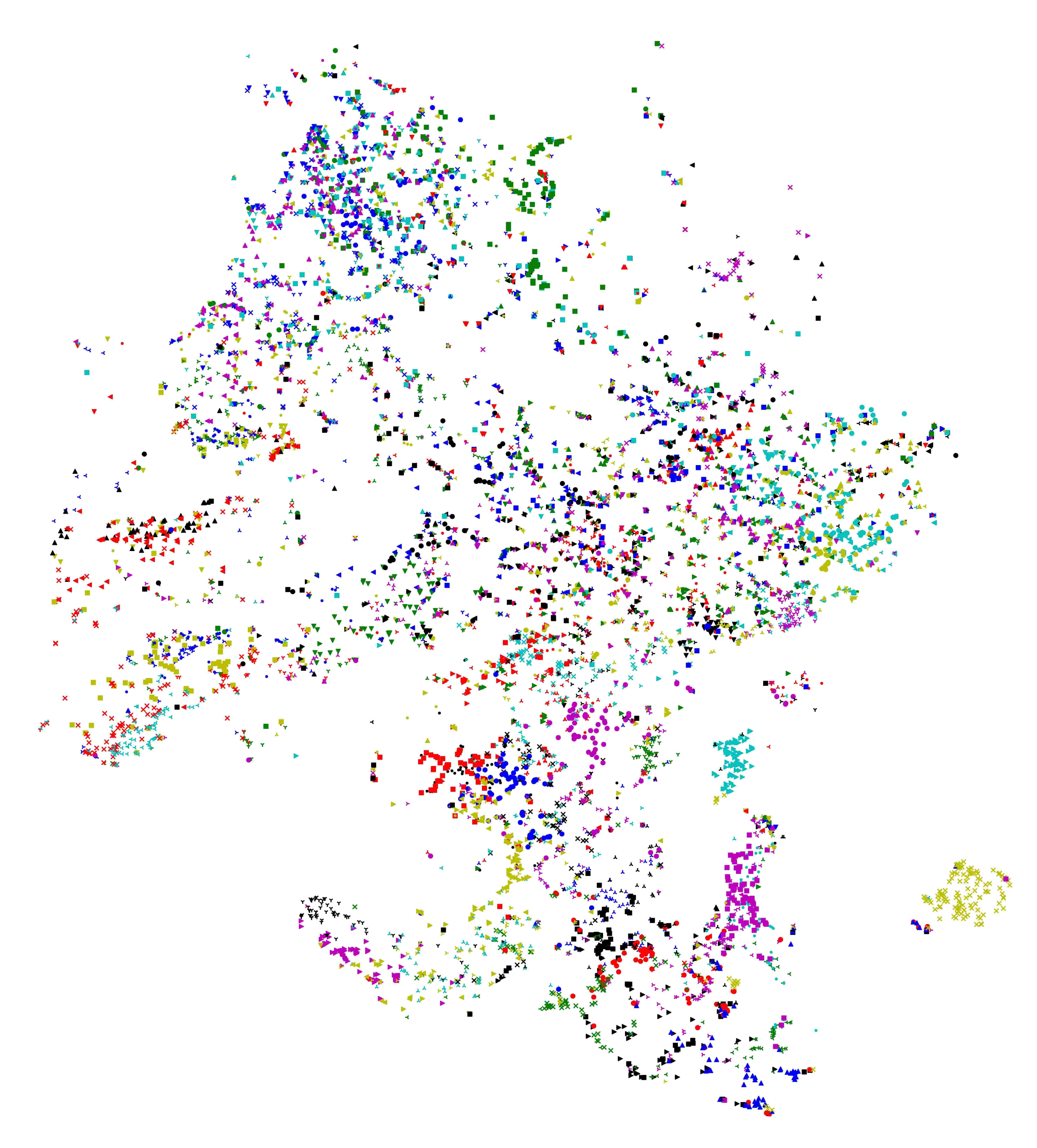} &
        \includegraphics[width=\fullsizeplotwidth]{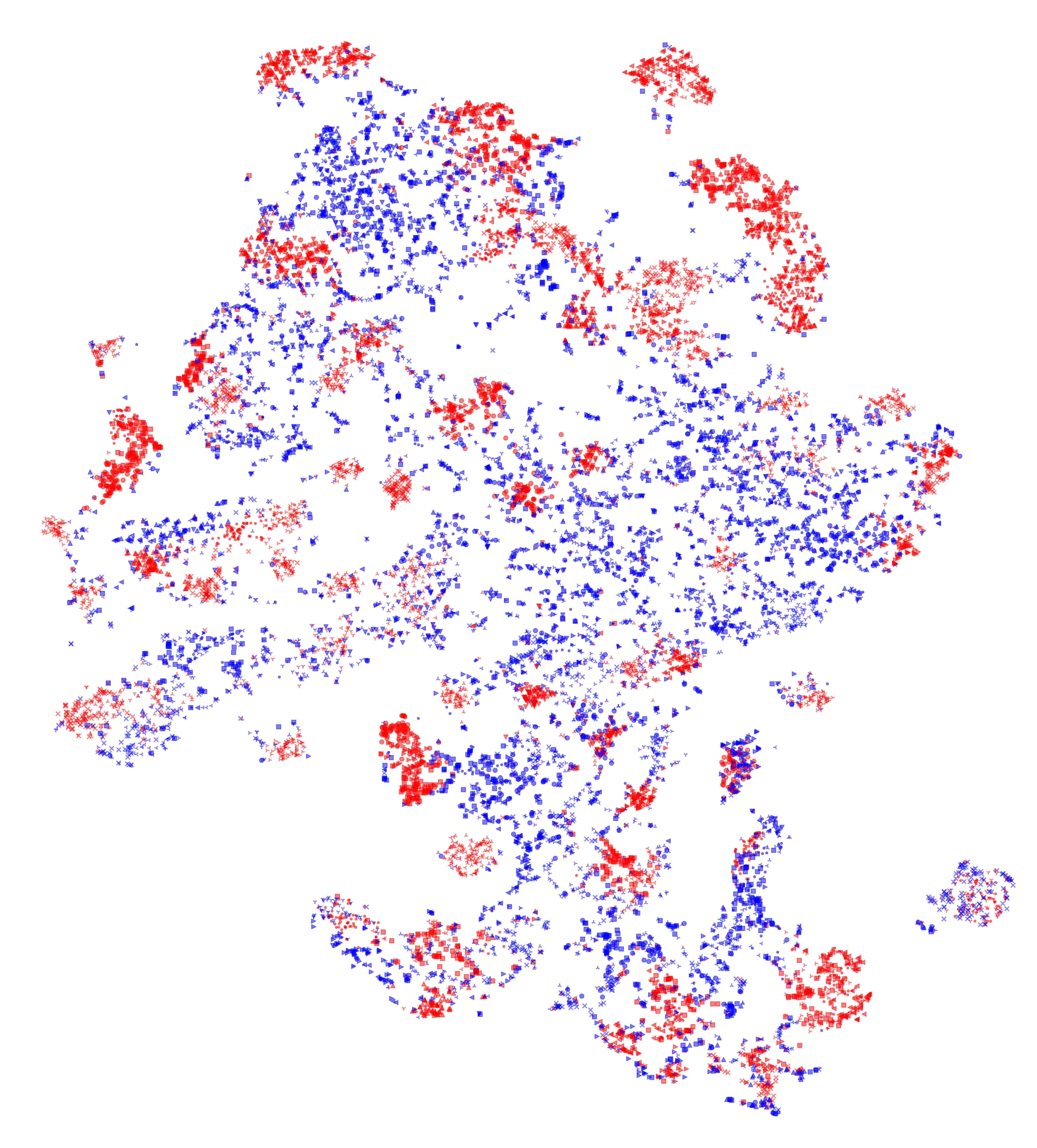} &
        \includegraphics[width=\fullsizeplotwidth]{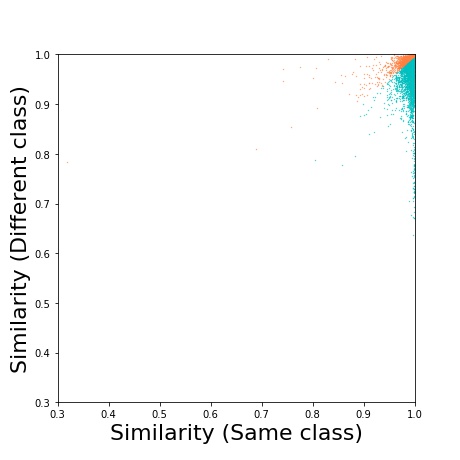} &
        \includegraphics[width=\fullsizeplotwidth]{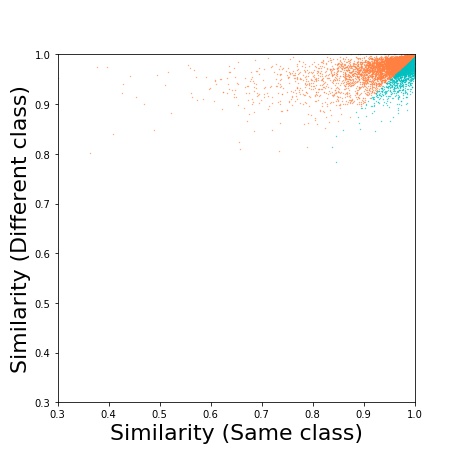}
        
        \\
        
        \raisebox{1cm}{\rotatebox{90}{N-pair}} &
        \includegraphics[width=\fullsizeplotwidth]{figs/tsne/fc/Npair/train.jpg} &
        \includegraphics[width=\fullsizeplotwidth]{figs/tsne/fc/Npair/test.jpg} &
        \includegraphics[width=\fullsizeplotwidth]{figs/tsne/fc/Npair/combined.jpg} &
        \includegraphics[width=\fullsizeplotwidth]{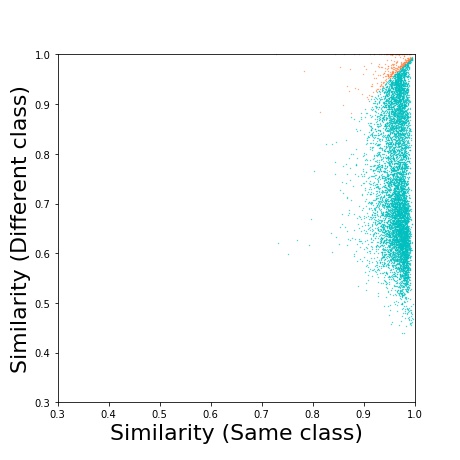} &
        \includegraphics[width=\fullsizeplotwidth]{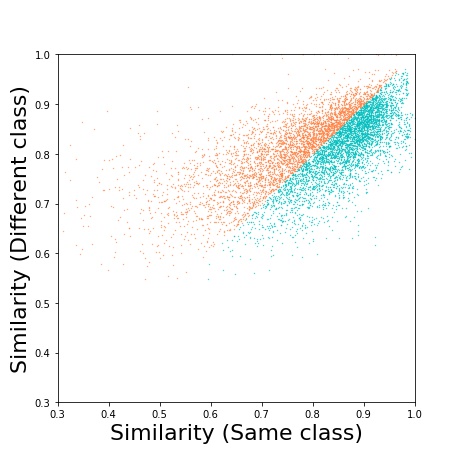}

        \\
        
        \raisebox{1.1cm}{\rotatebox{90}{SHN}} &
        \includegraphics[width=\fullsizeplotwidth]{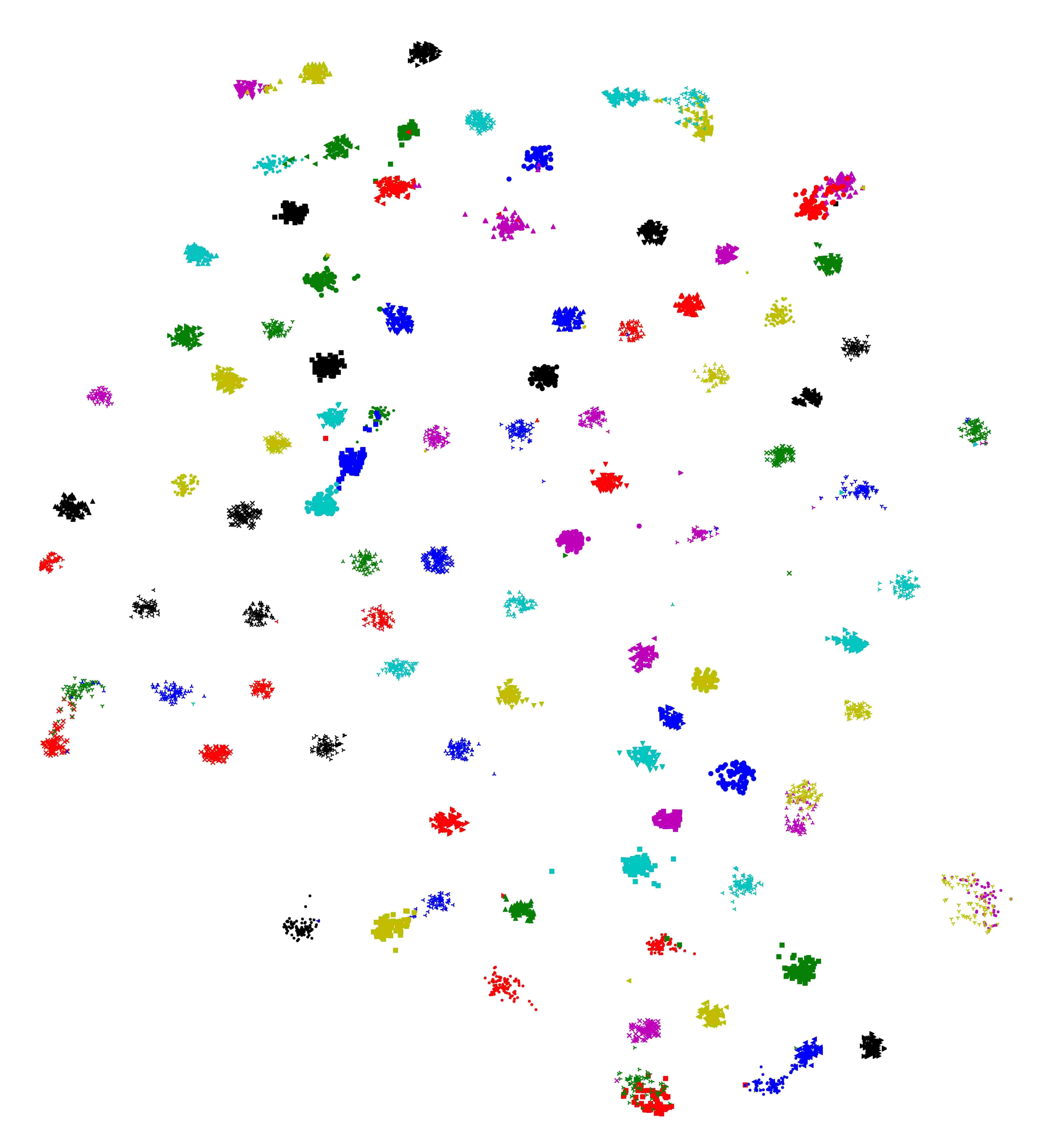} &
        \includegraphics[width=\fullsizeplotwidth]{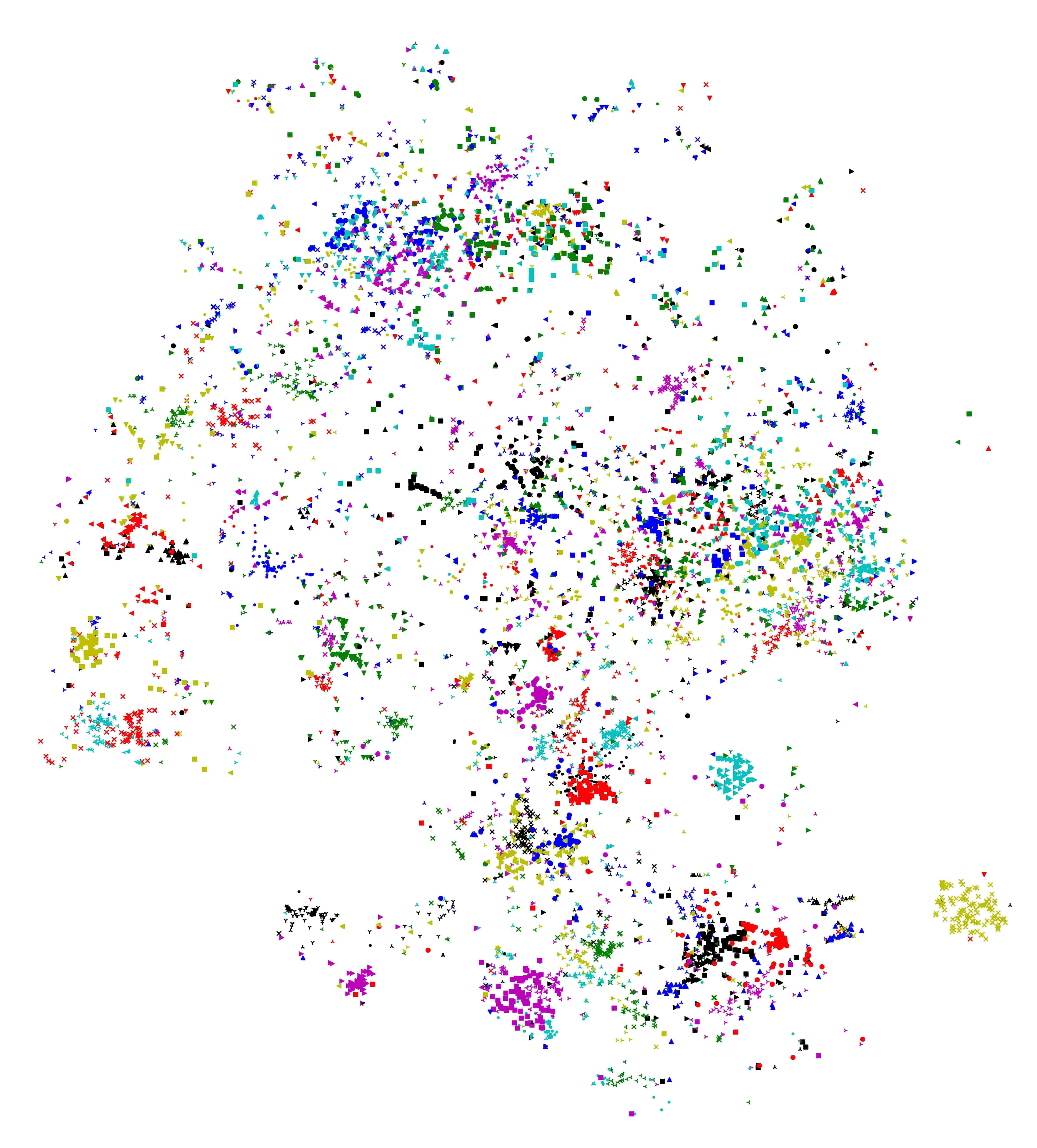} &
        \includegraphics[width=\fullsizeplotwidth]{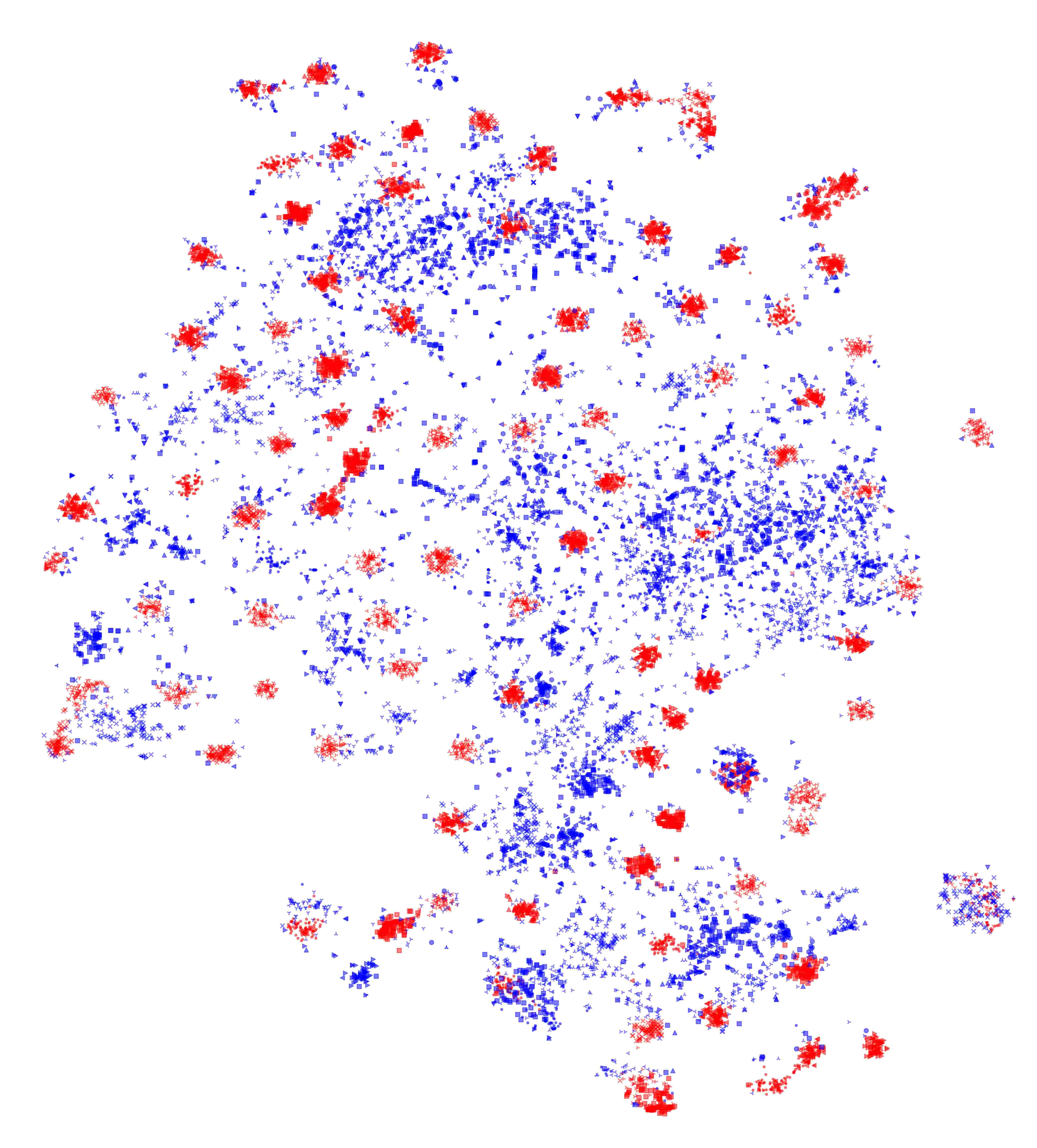} &
        \includegraphics[width=\fullsizeplotwidth]{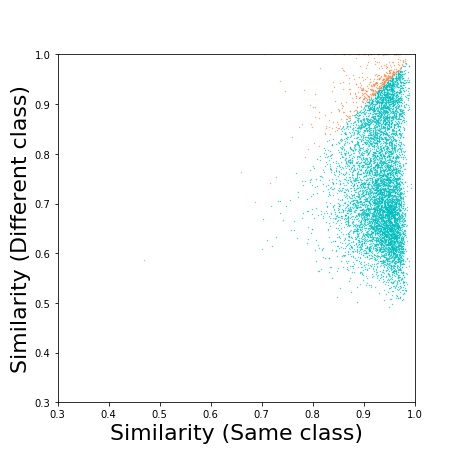} &
        \includegraphics[width=\fullsizeplotwidth]{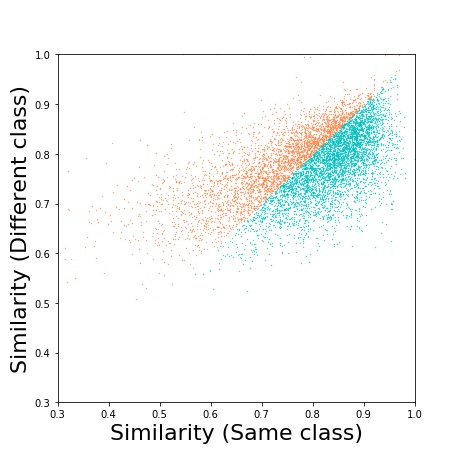}
        
        \\
        
        \raisebox{.8cm}{\rotatebox{90}{EPSHN}} &
        \includegraphics[width=\fullsizeplotwidth]{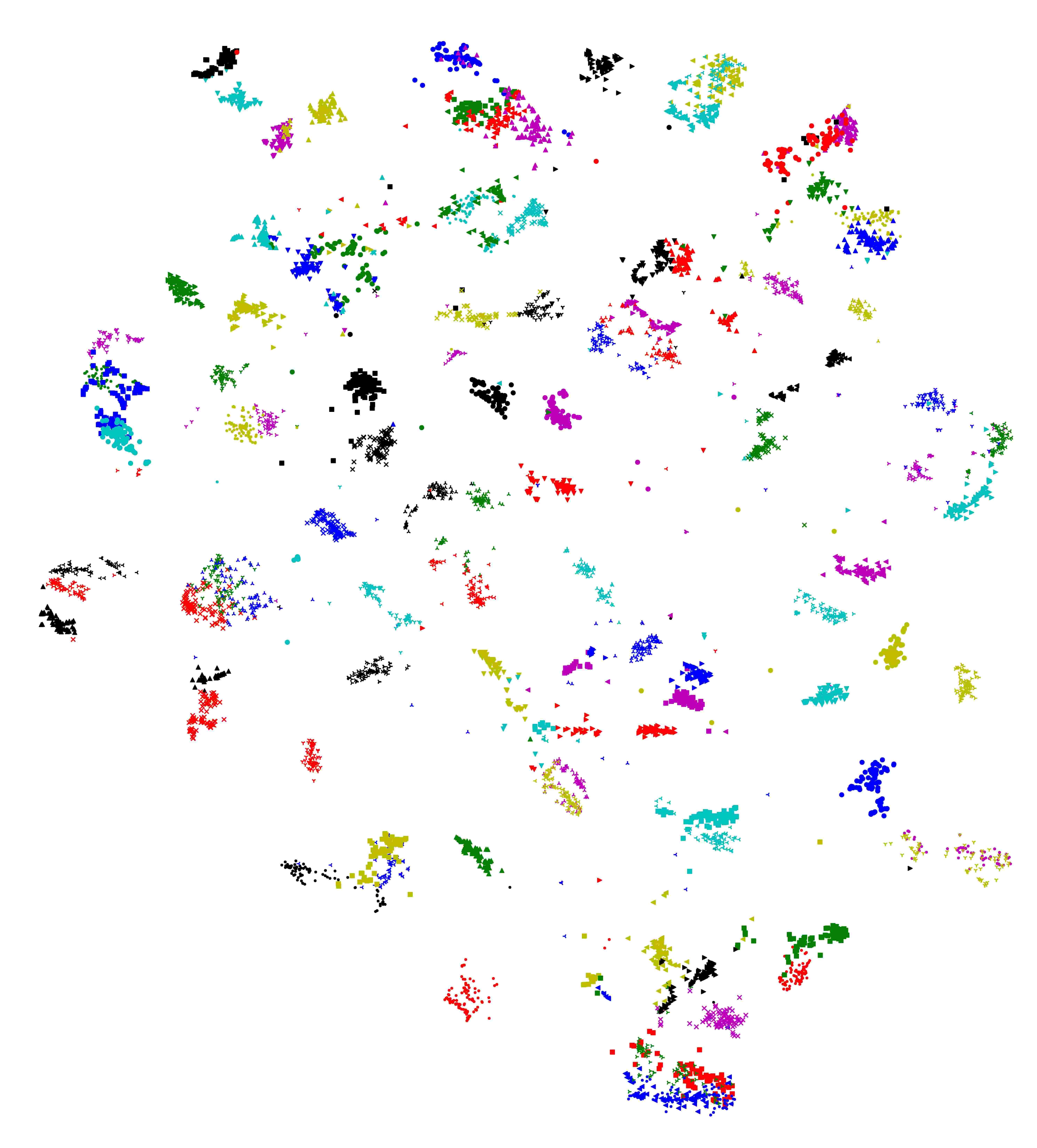} &
        \includegraphics[width=\fullsizeplotwidth]{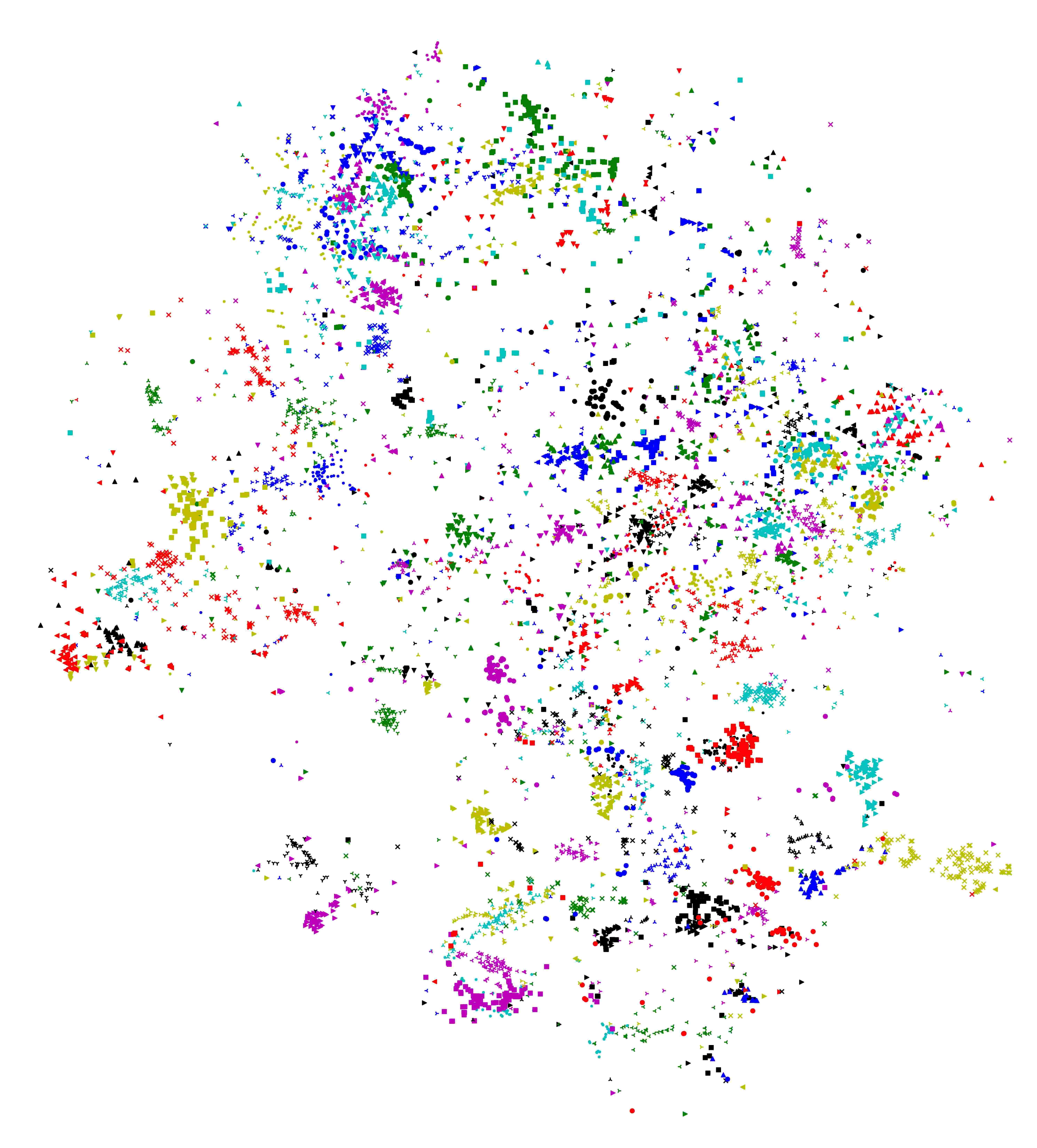} &
        \includegraphics[width=\fullsizeplotwidth]{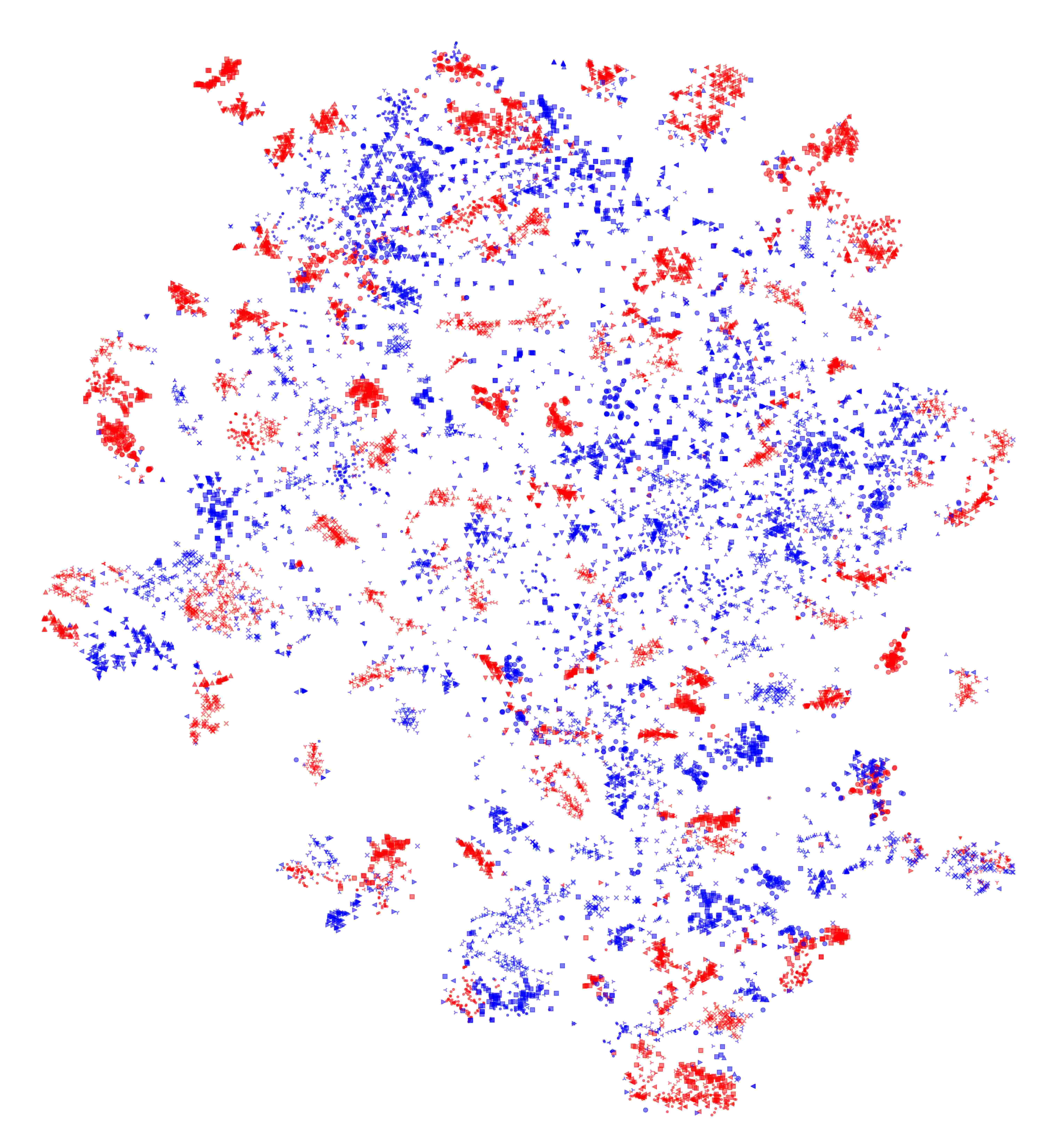} &
        \includegraphics[width=\fullsizeplotwidth]{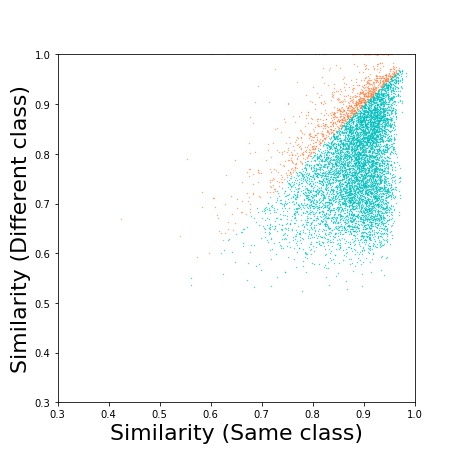} &
        \includegraphics[width=\fullsizeplotwidth]{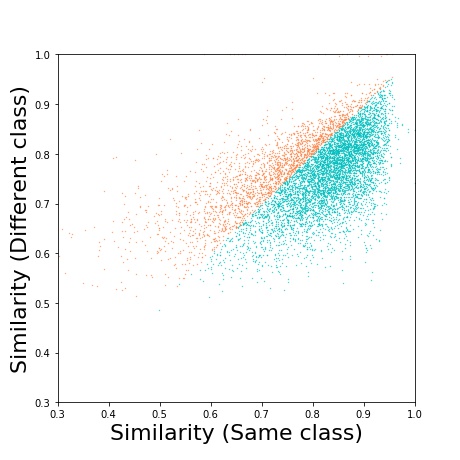}
    \end{tabular}
    \caption{For each embedding approach, we show the joint t-SNE embedding of training and test data from the CAR dataset~\cite{CAR196}, and both together (training data in red, testing data in blue); collectively, these show how the test data is embedded relative to the training data. The scatter plots showing the distance to the most similar example from the same and different classes in training and testing data. Examples that are closer to a same class result are colored in cyan, while examples that are closer to a different class result are color in orange.}
\label{fig:full_page}
\end{figure*}

\section{Discussion}
Table~\ref{table:CAR} shows the Recall@1 accuracy for each of the different triplet selection approaches on the CAR~\cite{CAR196} dataset. Figure~\ref{fig:full_page} shows each of our visualizations for the different approaches. In this section we share what these visualizations tell us. 

\begin{figure*}
    \centering
    \begin{subfigure}[b]{.51\textwidth}
        \includegraphics[width=\columnwidth]{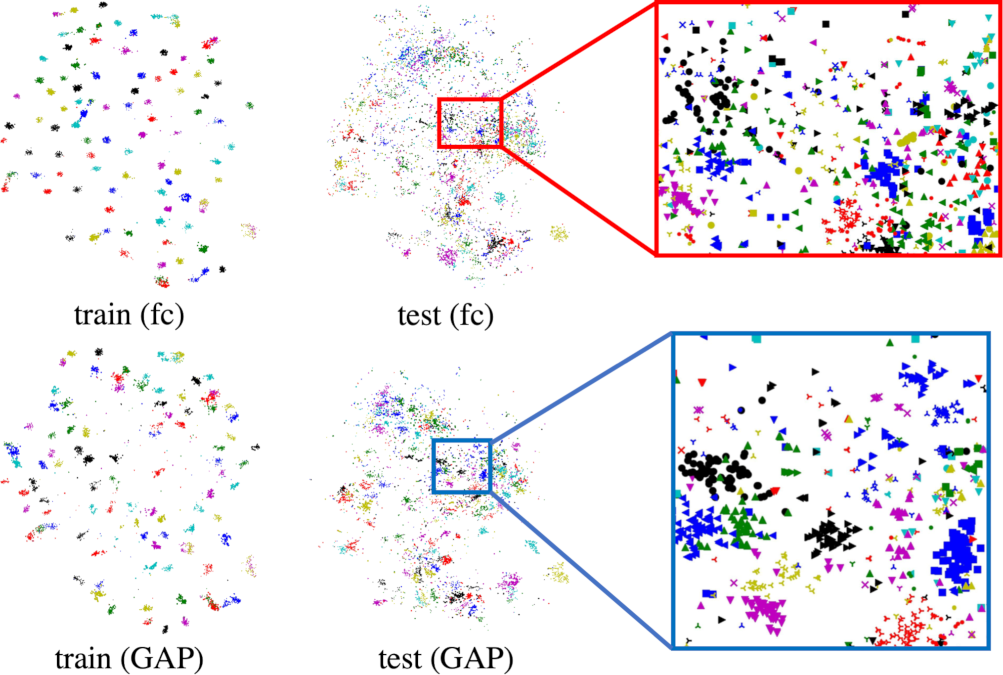}
        \caption{}
    \end{subfigure}
    \hspace{.05cm}
    \begin{subfigure}[b]{.36\textwidth}
        \includegraphics[width=.49\columnwidth]{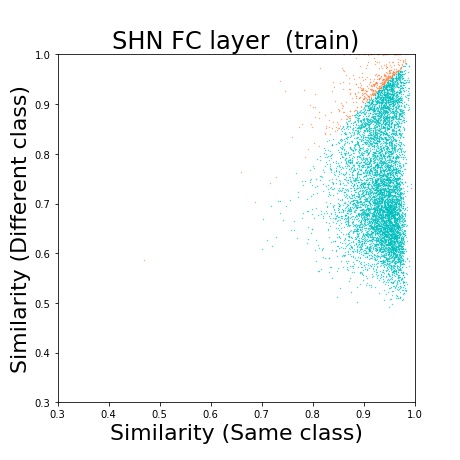}\includegraphics[width=.49\columnwidth]{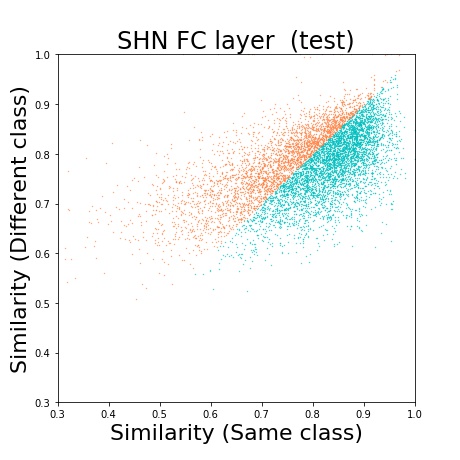}
        \\
        \includegraphics[width=.49\columnwidth]{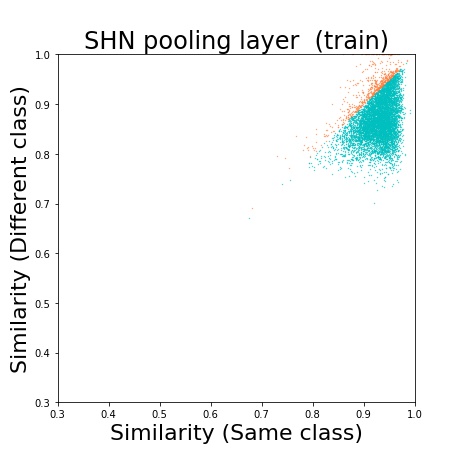}\includegraphics[width=.49\columnwidth]{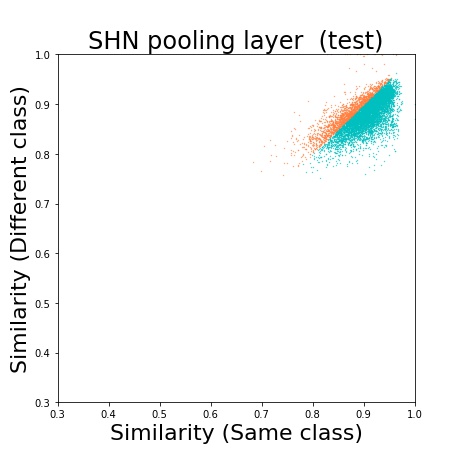}
        \caption{}
    \end{subfigure}
	 \caption{(a) t-SNE visualization of Semi-Hard Negative on CAR training and testing data, using the embedding from the fully connected (FC) layer (top), and from the Global Average Pooling (GAP) layer (bottom).  The embedding of the training and test data are shown along with a zoomed in region highlighting differences in the structure of the aligned embeddings.  The scatter plots in (b) show the similarity to the closest same-class and different-class images. This highlights that the performance drops much less from training to testing data using the GAP embedding.}
	 \label{fig:fc_vs_gp}
\end{figure*}

\paragraph{Batch All} The t-SNE of the training data shows that the representation clusters the training classes, but often multiple classes partially overlap. The testing data shows a few clear clusters, but is much more scattered, and the joint embedding picture shows that the test data is often mapped to the same location as the training data. The closest same class vs. closest different class similarity scatter plots show that most images have a very similar image from both the same category and from a different category. For testing data, the similarity to both similar and different classes remains quite high, but more often images are similar to more different classes, a reflection of the poor R@1 accuracy. 

\paragraph{N-Pairs} The t-SNE shows the training data is very well clustered, which is consistent with the scatter plots showing that most points are very similar to their closest same class image, but not usually as similar to any different class image. However, on testing data, the t-SNE shows the new classes to be partially clustered, with classes generally localized but overlapping other new classes. The joint embedding shows new points sometimes overlapping the existing classes and sometimes forming clusters away from existing classes. The scatter plot shows the similarity to the most similar image is less in the testing than the training data, and that the most similar image is more often from a different class. 

\paragraph{Semi-Hard Negative (SHN)} The t-SNE plots again show strong clustering for the training data, and on testing data, perhaps slightly improved clustering relative to N-pairs (e.g. the clusters along the left edge), and a similar overlap between the embedding of the training and testing data. The scatter plots show the closest same class similarity for the training data is very high, but decreases substantially for test data. 

\paragraph{Easy Positive Semi-Hard Negative (EPSHN)} The t-SNE plots for the training data show strong clusters, but those clusters have more structure because only the most similar pairs are drawn together (instead of random pairs or all pairs). The test data continue to show strong clustering, and when overlapped, the test classes are rarely embedding on top of the training classes. The scatter plots show that between closes same class images, the similarity is not as high as other approaches, but the behavior remains more similar on test data, suggesting that the representation generalizes well. 

\paragraph{Fully Connected vs Global Average Pooling Features}
A recent paper~\cite{vo2019generalization} shows that metric learning approaches generalize substantially better when using the Global Average Pooling (GAP) layer~\cite{vo2019generalization} instead of the final fully connected (FC) layer. We can see this in Table~\ref{table:CAR}, and in Figure~\ref{fig:fc_vs_gp}, we compare the visualizations for the FC and GAP embeddings for the Semi-Hard Negative approach, which is most improved by using the GAP layer. In the t-SNE embeddings, we can see that the GAP features are less tightly clustered than the FC features on the training data, but much more clearly clustered on the test data. This is especially apparent in the zoomed in regions showing roughly the same data points more tightly clustered for the GAP embedding. For the FC layer, the training data is very well clustered (all same class similarity is high, and different class similarity is often lower), but the GAP layer on training data, while usually still correct, has more cases where the different class similarity is almost as high. However, on testing data, the GAP layer retains the ability to give the closest images from the same (new) classes a higher similarity than the closest image from a different class.

The scatter plots also show that the GAP layer always has higher similarity (all the points are shifted up and to the right). This is because in the Resnet18 architecture the GAP features come from a ReLU layer, so are all positive (unlike the FC features where each element can be positive or negative).

\section{Conclusion}
Many approaches to image embedding have been proposed recently, and they are typically compared based on performance across a collection of datasets. Here we try to give new approaches to understanding what each embedding is doing, and specifically how the embedding generalizes to new data. The literature highlights that embedding approaches such as Semi-Hard Negative and Easy Positive Semi-Hard Negative, which use fewer triplets in a batch and impose fewer constraints, often outperform approaches that use more triplets, like N-Pairs and Batch All, on testing data. Our visualization tools show that these approaches with fewer triplets often do not cluster the training data as well, but the embedding better preserves the similarity of new data.

To show how this generalizes to other datasets, the appendix contains a duplicate of Figure~\ref{fig:full_page} for the CUB dataset~\cite{CUB200}. Additionally, to support replication of results, all code to re-generate the figures in this paper, or generate these visualizations for othe datasets will be made public.

\nocite{langley00}

\clearpage 

\bibliography{example_paper}
\bibliographystyle{icml2019}

\appendix
\onecolumn
\section*{Appendix}
This appendix replicates the figures for the CUB dataset. All network architecture and optimization choices are the same as the main text.

\begin{table*}[h]
\setlength{\tabcolsep}{0.4em}
\begin{center}
\begin{tabular}{|c|c|c|c|c|}
\hline
Method  & BatchAll & Npair & SHN & EPSHN \\
\hline
FC (train) 
 & 90.91 & 92.18 & 86.72 & 85.79\\
GAP (train) 
 & 89.70 & 87.57 & 82.33 & 81.39\\
FC (test) 
 & 38.86 & 50.85 & 53.55 & 56.23\\
GAP (test) 
 & 59.70 & 63.23 & 63.32 & 63.51\\
\hline
\end{tabular}
\end{center}
\caption{Recall@1 Performance on the CUB dataset}
\label{table:CUB}
\end{table*}

\setlength{\tabcolsep}{1pt}
\renewcommand{\arraystretch}{.5}
\begin{figure*}[h]
    \centering
    \begin{tabular}{ccccccc}
        \centering
        & \parbox{1.5cm}{\raggedright \centering t-SNE\\ (train)}
        & \parbox{1.5cm}{\raggedright \centering t-SNE\\ (test)} 
        & \parbox{1.5cm}{\raggedright \centering t-SNE\\ (combined)} 
        & \parbox{1.5cm}{\raggedright \centering scatter\\ (train)} 
        & \parbox{1.5cm}{\raggedright \centering scatter\\ (test)}\\
        
        \raisebox{.7cm}{\rotatebox{90}{Batch-All}} & 
        \includegraphics[width=\fullsizeplotwidth]{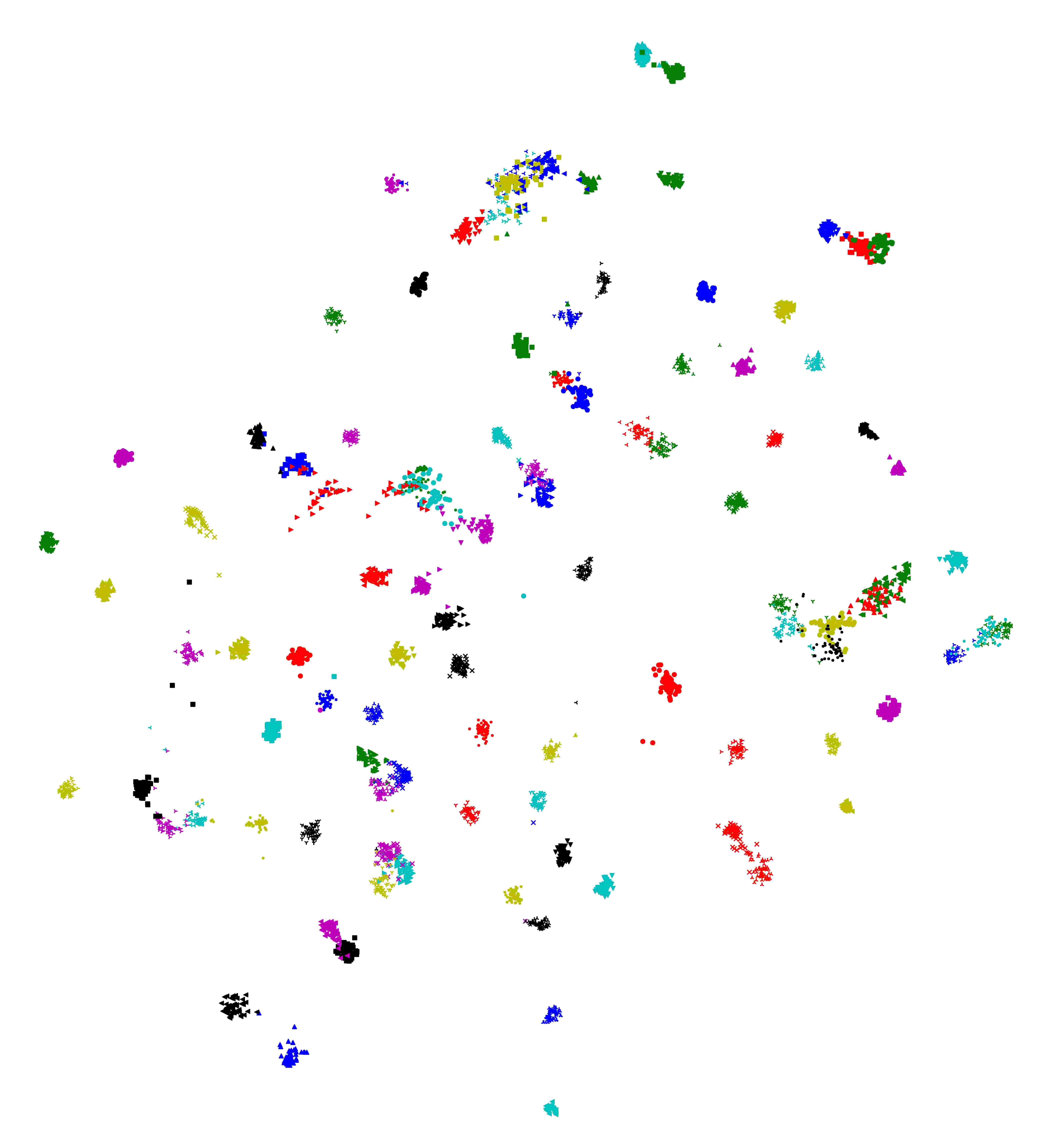} &
        \includegraphics[width=\fullsizeplotwidth]{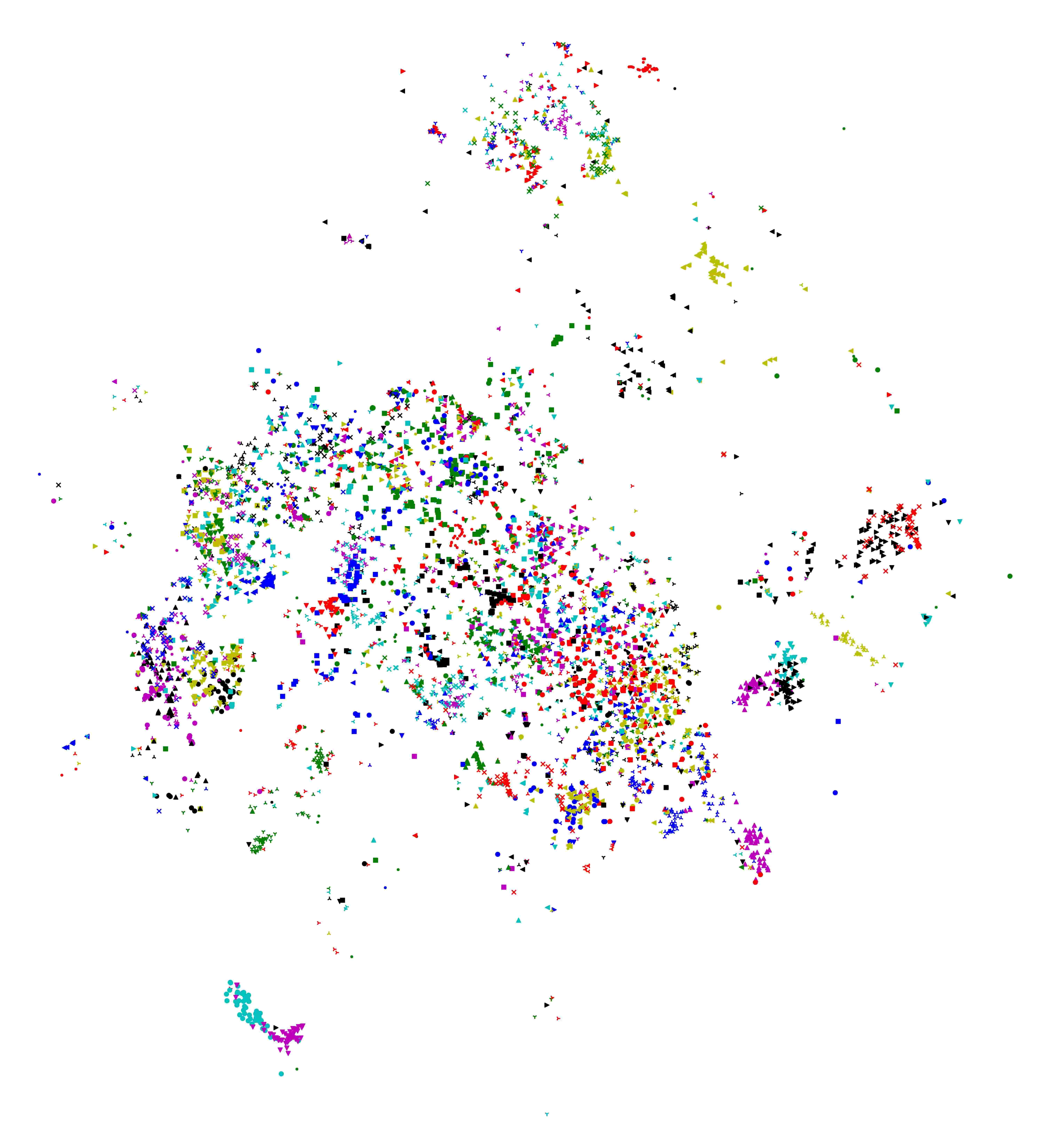} &
        \includegraphics[width=\fullsizeplotwidth]{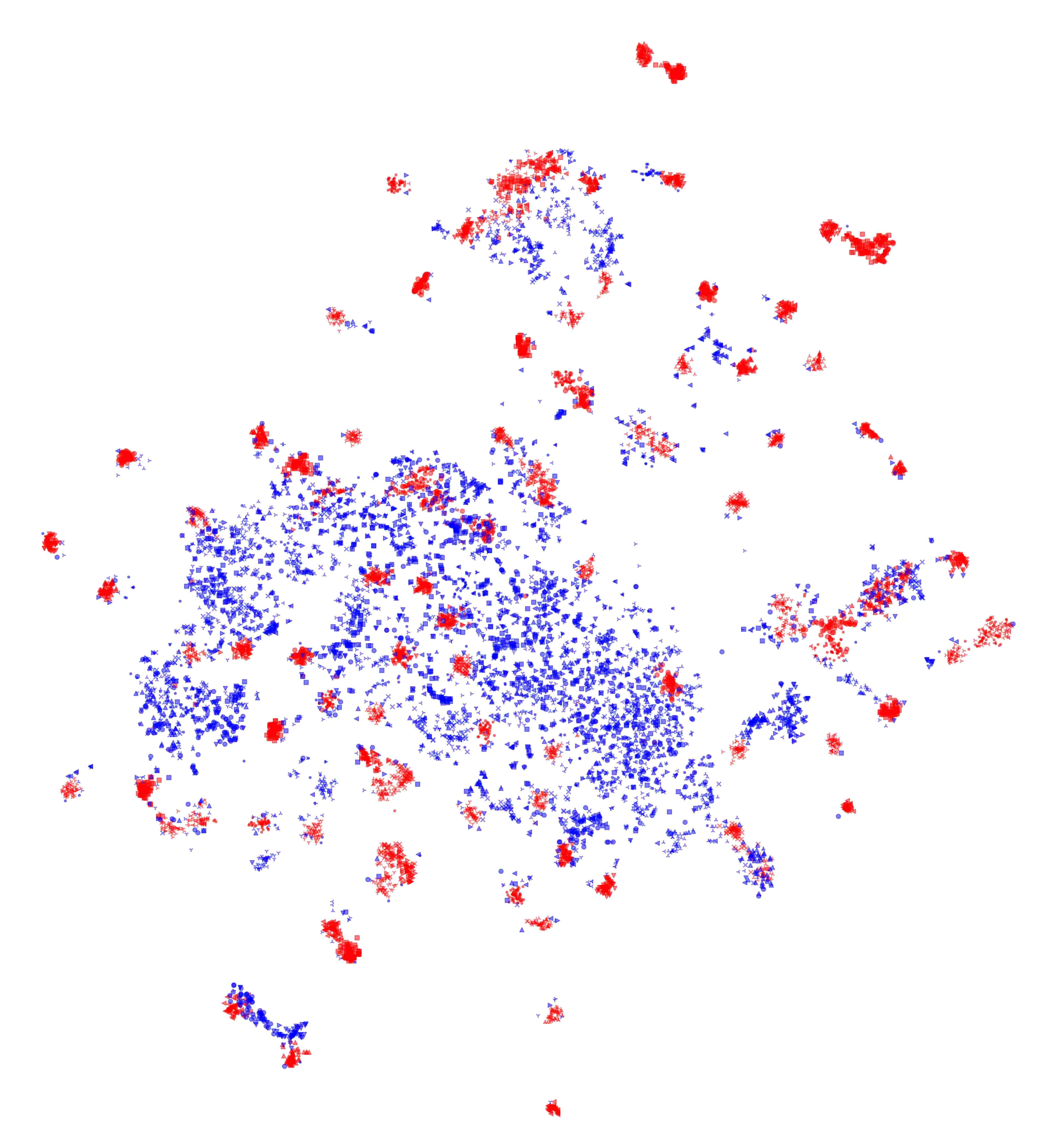} &
        \includegraphics[width=\fullsizeplotwidth]{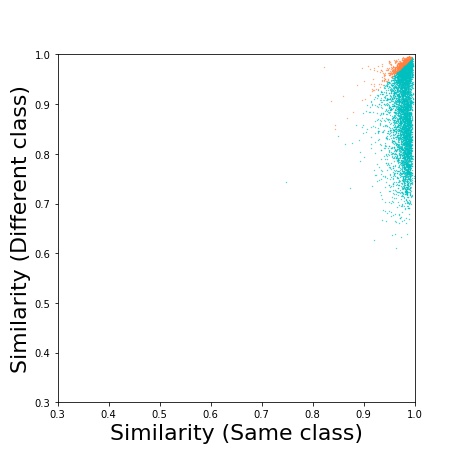} &
        \includegraphics[width=\fullsizeplotwidth]{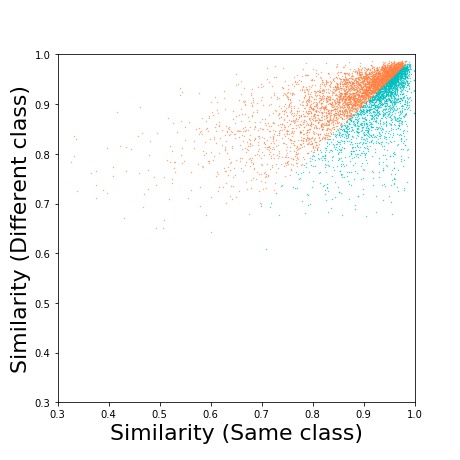}
        
        \\
        
        \raisebox{1cm}{\rotatebox{90}{N-pair}} &
        \includegraphics[width=\fullsizeplotwidth]{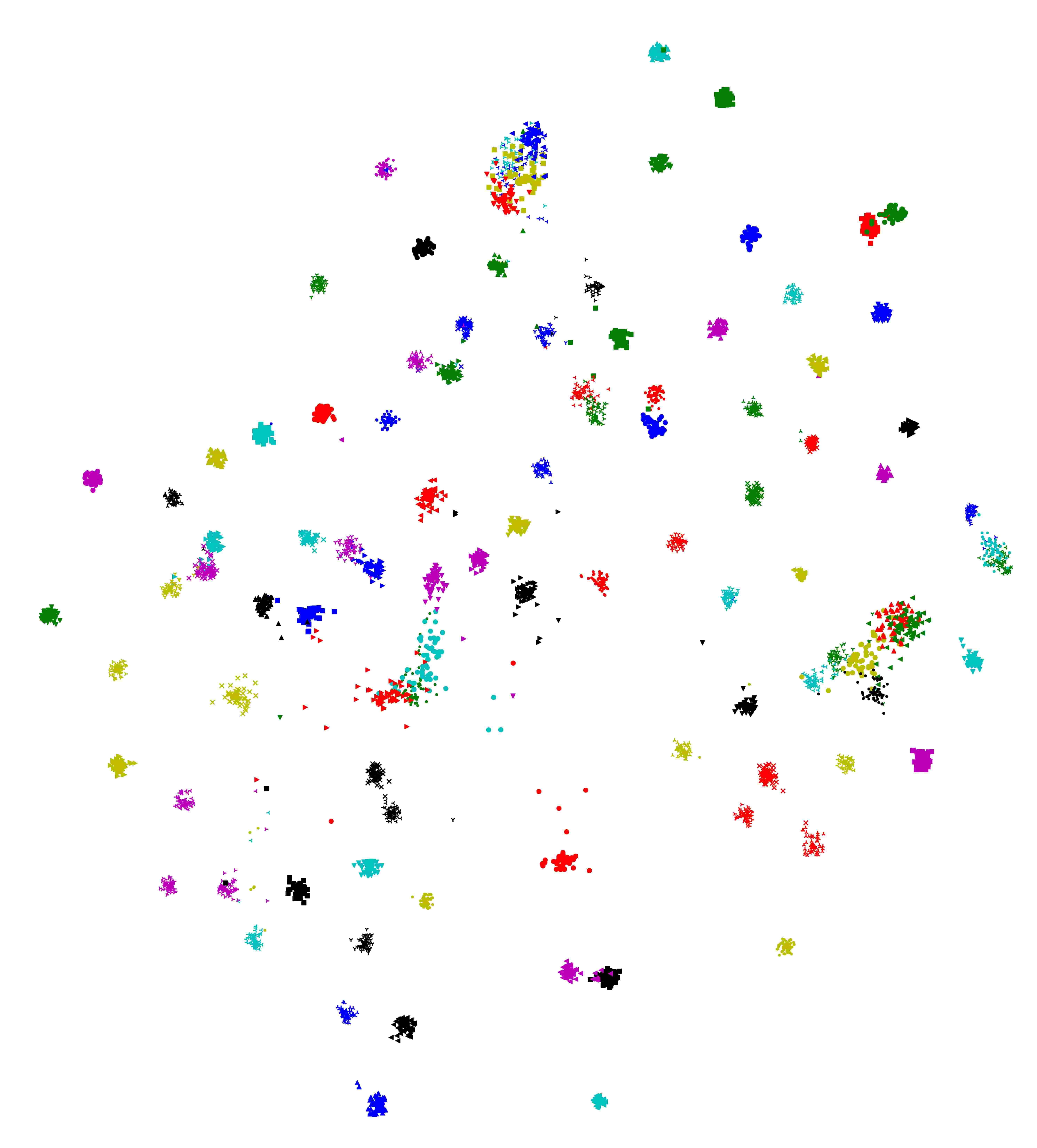} &
        \includegraphics[width=\fullsizeplotwidth]{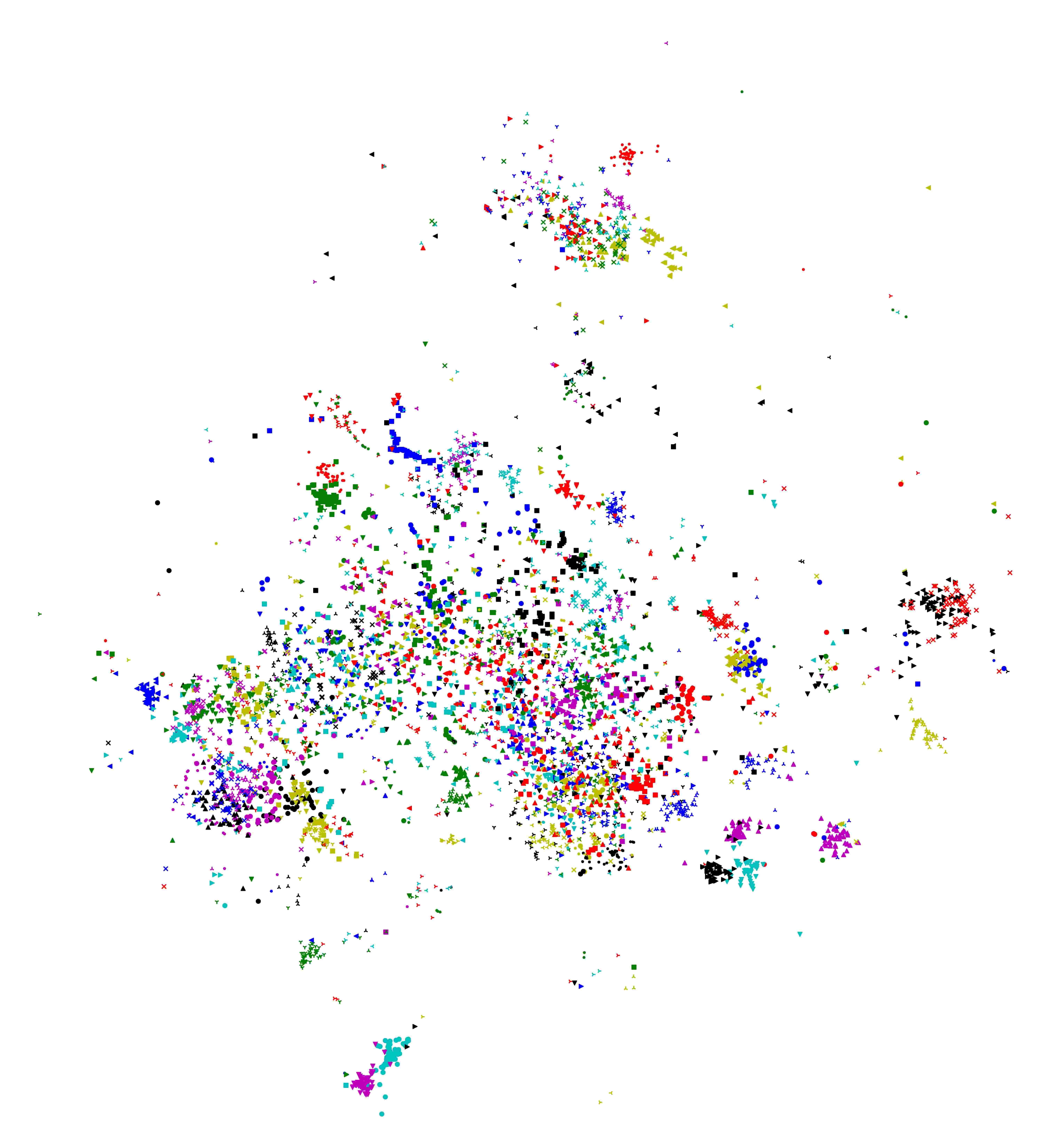} &
        \includegraphics[width=\fullsizeplotwidth]{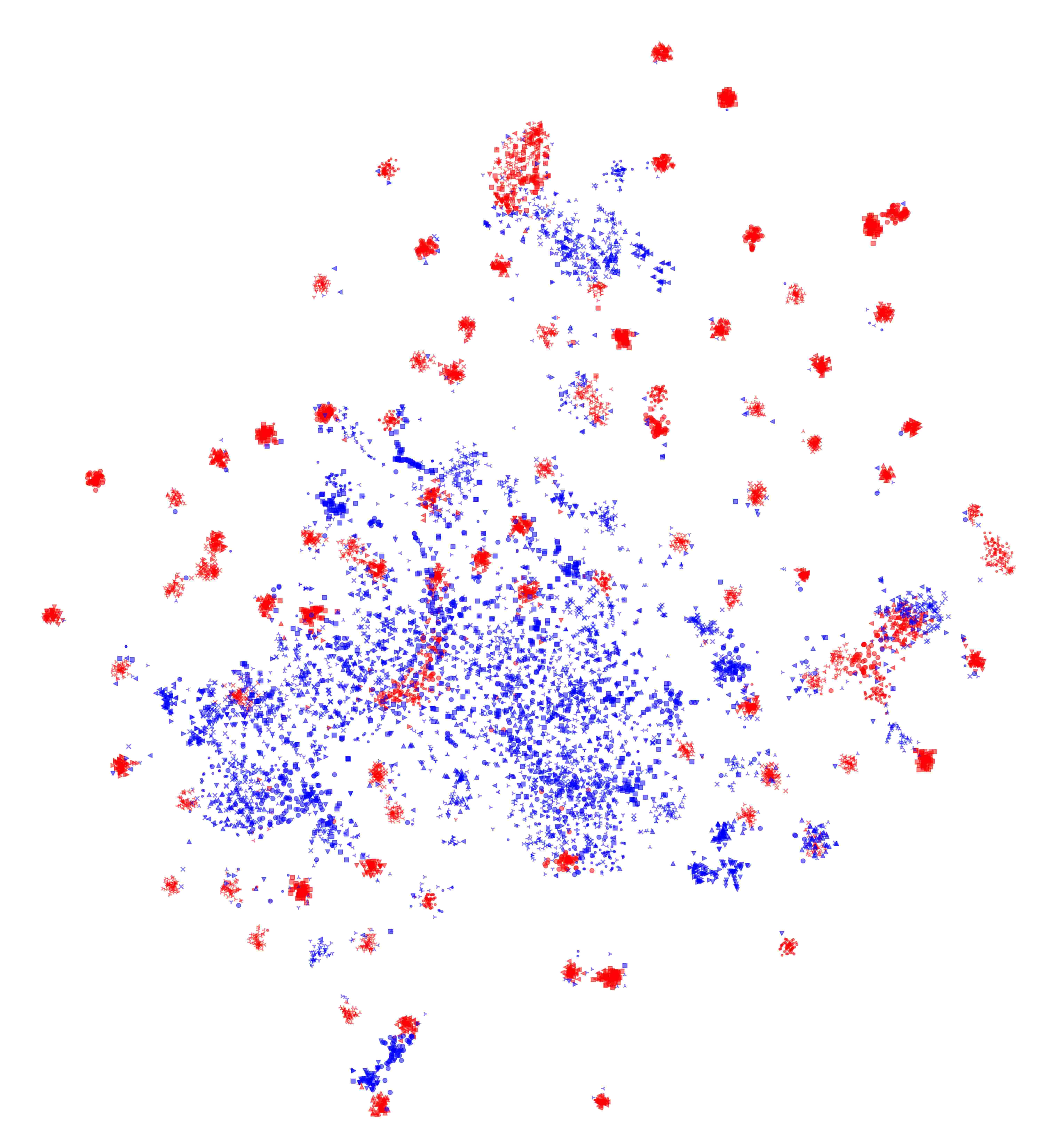} &
        \includegraphics[width=\fullsizeplotwidth]{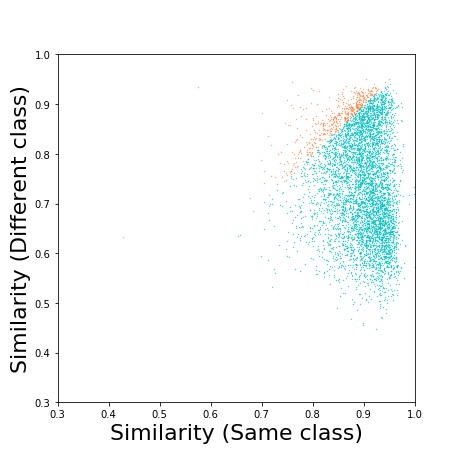} &
        \includegraphics[width=\fullsizeplotwidth]{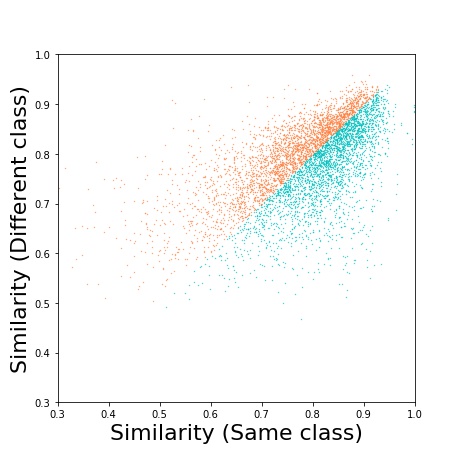}

        \\
        
        \raisebox{1.1cm}{\rotatebox{90}{SHN}} &
        \includegraphics[width=\fullsizeplotwidth]{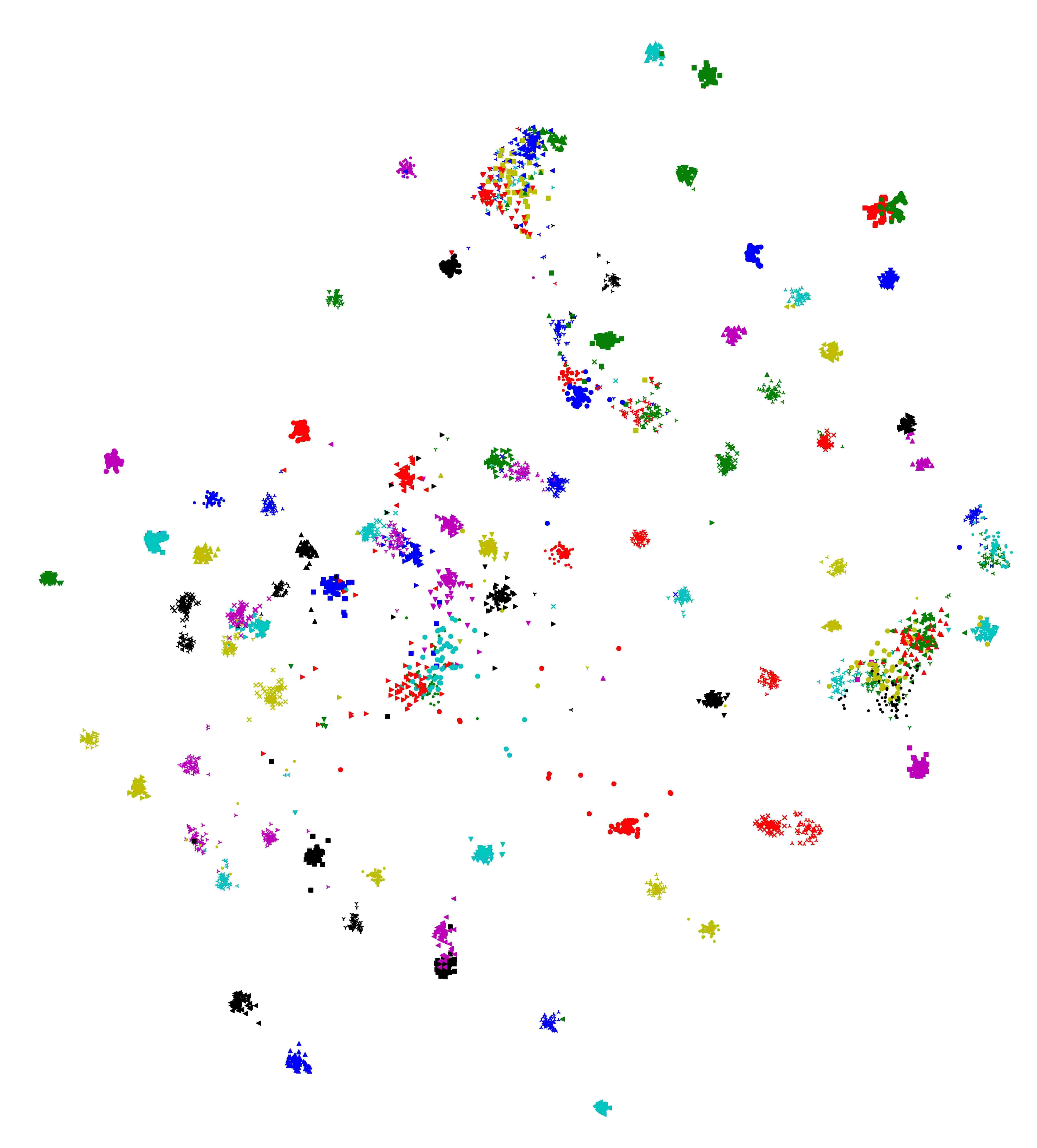} &
        \includegraphics[width=\fullsizeplotwidth]{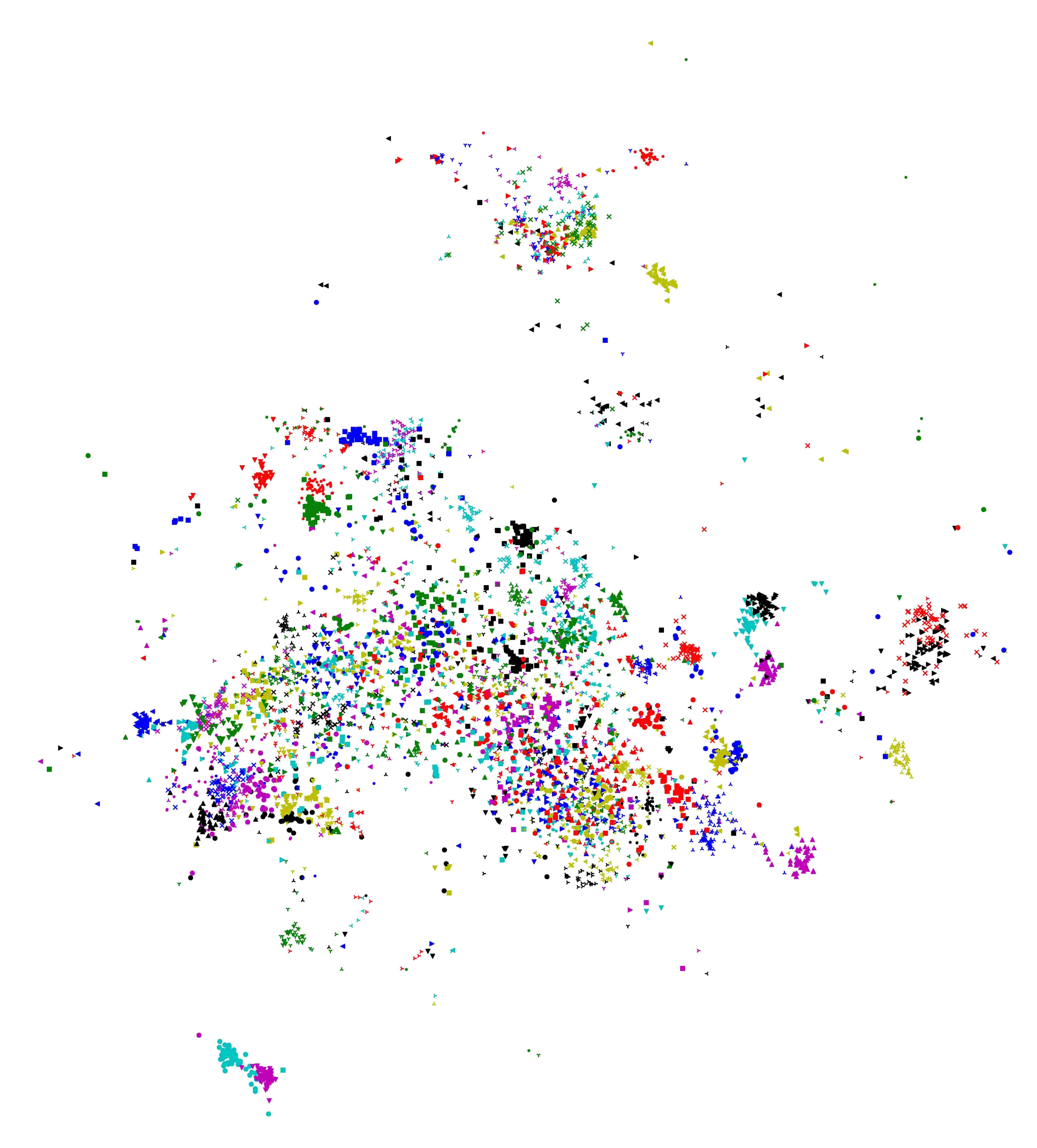} &
        \includegraphics[width=\fullsizeplotwidth]{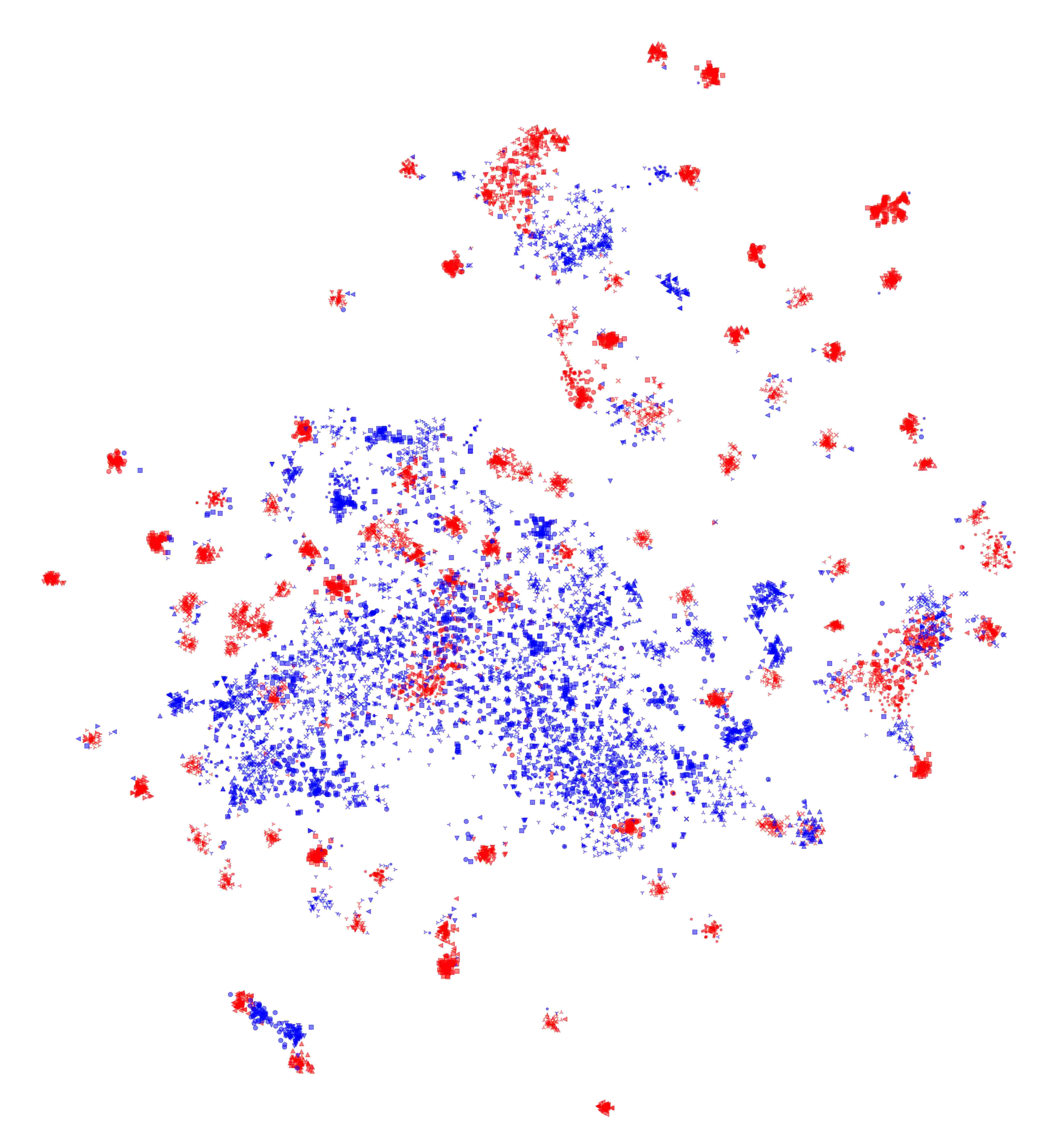} &
        \includegraphics[width=\fullsizeplotwidth]{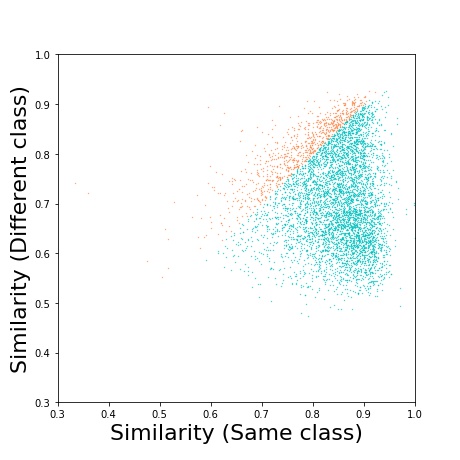} &
        \includegraphics[width=\fullsizeplotwidth]{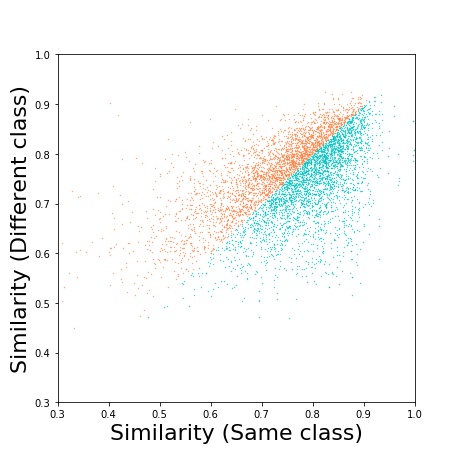}
        
        \\
        
        \raisebox{.8cm}{\rotatebox{90}{EPSHN}} &
        \includegraphics[width=\fullsizeplotwidth]{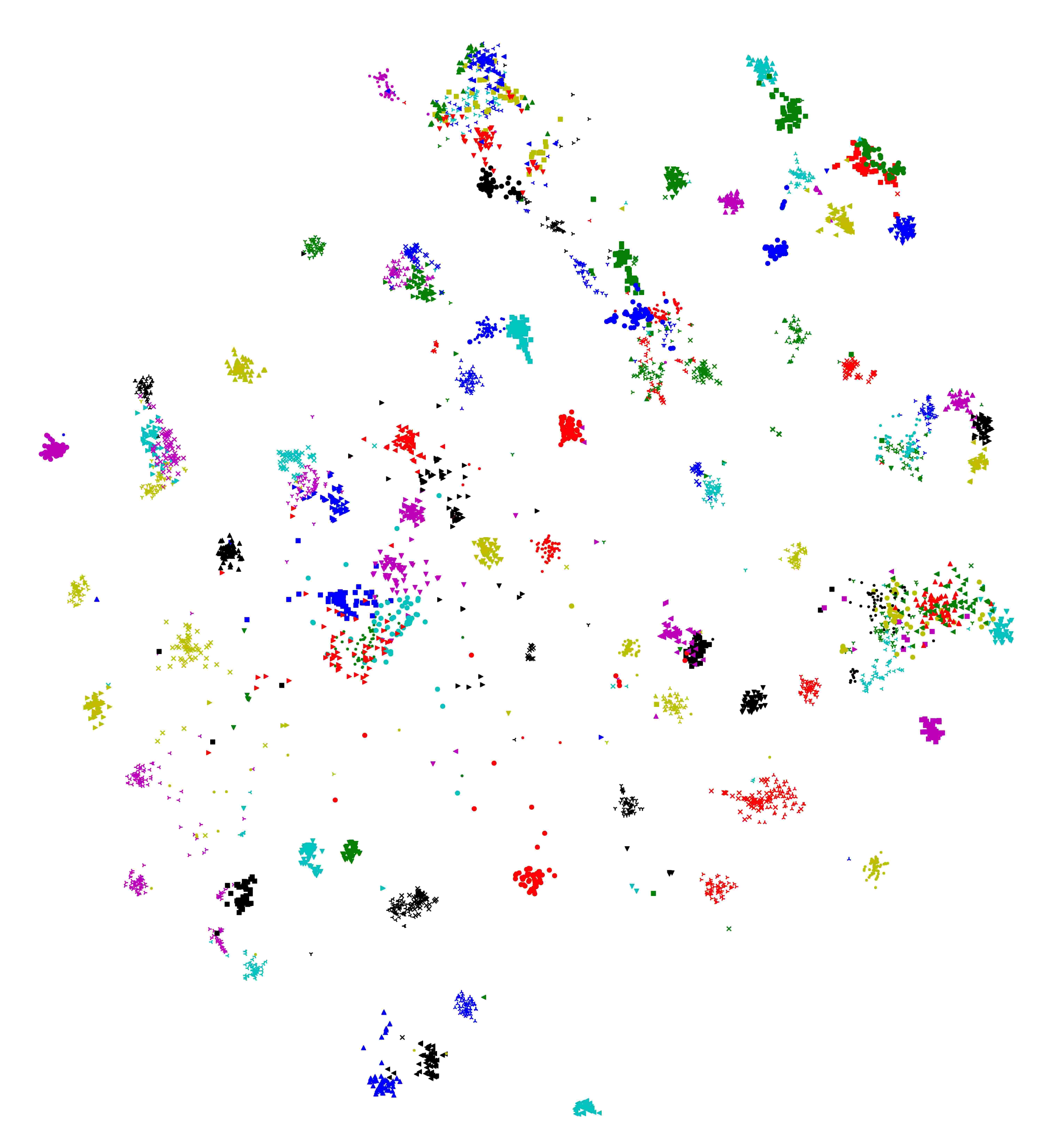} &
        \includegraphics[width=\fullsizeplotwidth]{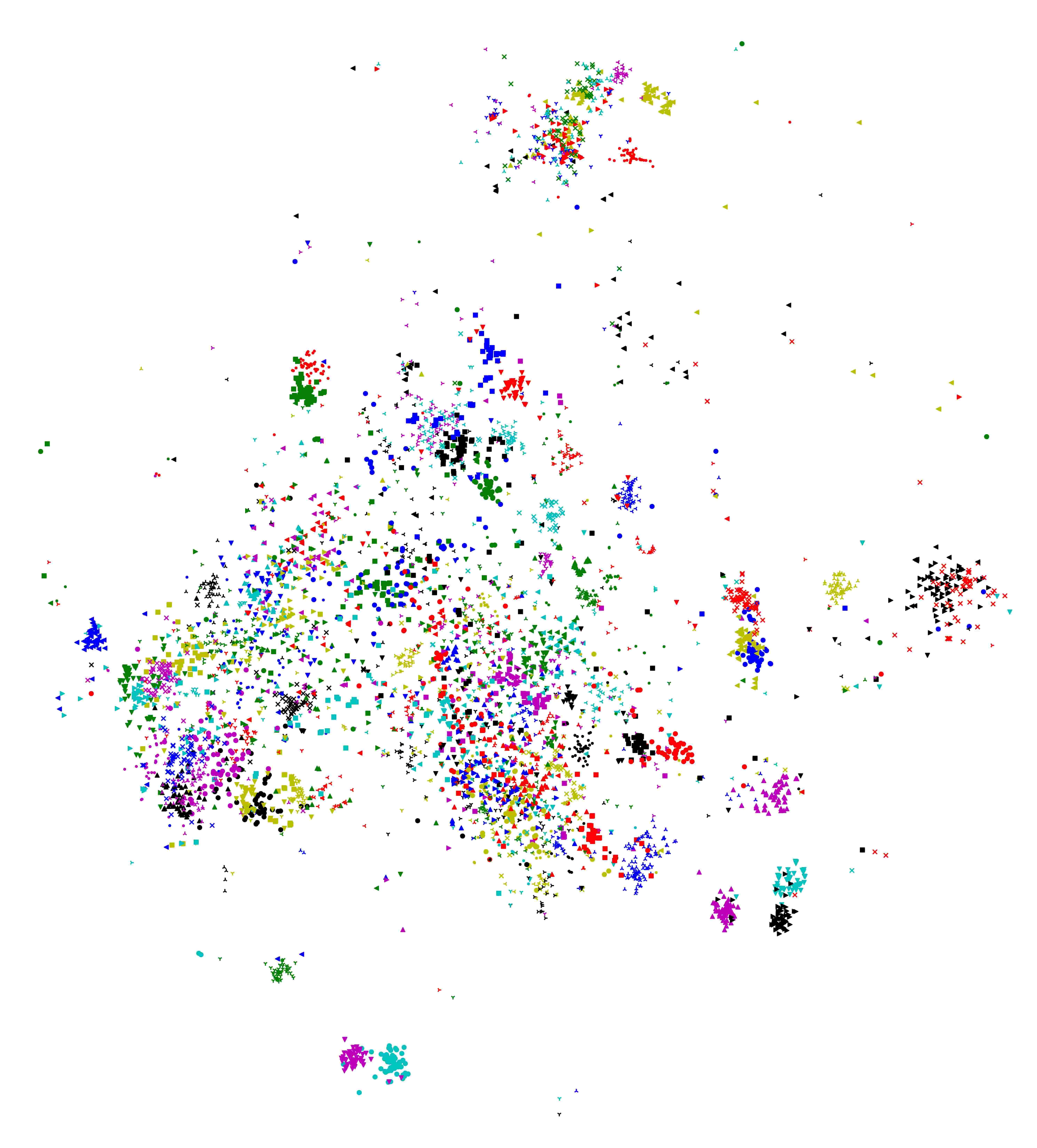} &
        \includegraphics[width=\fullsizeplotwidth]{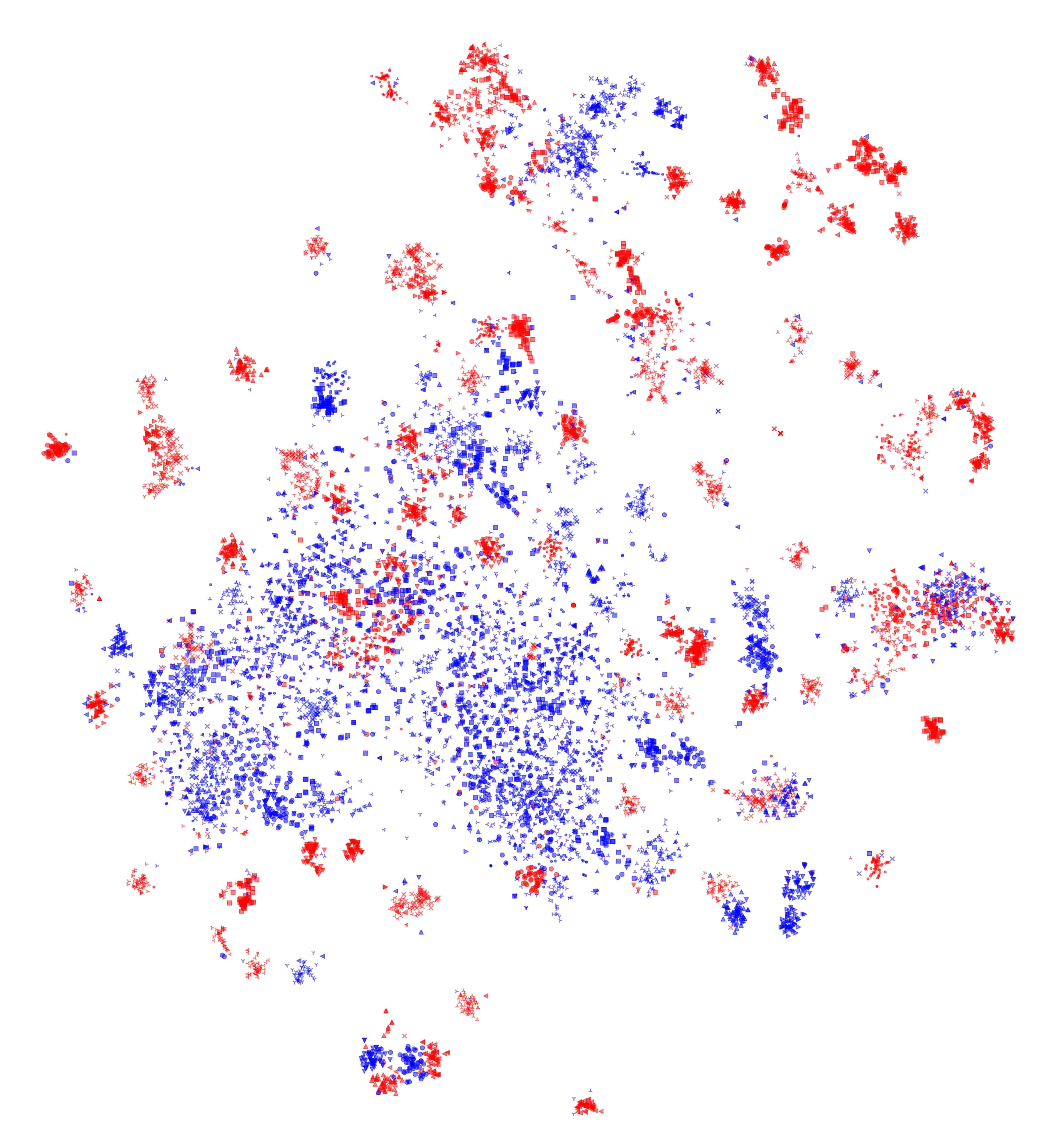} &
        \includegraphics[width=\fullsizeplotwidth]{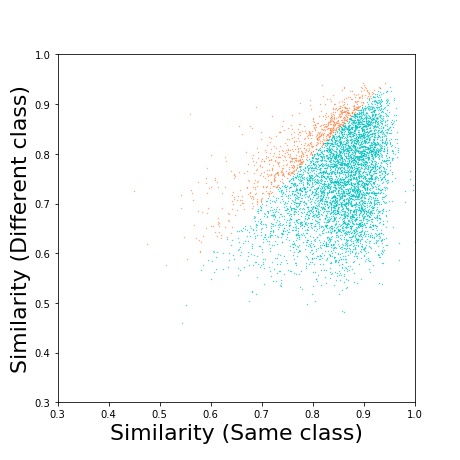} &
        \includegraphics[width=\fullsizeplotwidth]{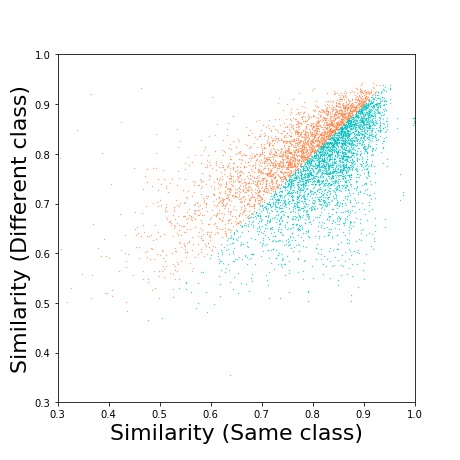}
    \end{tabular}
    \caption{For each embedding approach, we show the joint t-SNE embedding of training and test data from the CUB dataset~\cite{CUB200}, and both together (training data in red, testing data in blue); collectively, these show how the test data is embedded relative to the training data. The scatter plots showing the distance to the most similar example from the same and different classes in training and testing data. Examples that are closer to a same class result are colored in cyan, while examples that are closer to a different class result are color in orange.}
\label{fig:full_page}
\end{figure*}

Code is released on:

\url{https://github.com/GWUvision/Embedding_visualization}

\end{document}